\documentclass{article} 
\usepackage{iclr2024_conference,times}

\usepackage{amsmath,amsfonts,bm}

\def\eqref#1{equation~\ref{#1}}

\def\1{\bm{1}}

\DeclareMathAlphabet{\mathsfit}{\encodingdefault}{\sfdefault}{m}{sl}
\SetMathAlphabet{\mathsfit}{bold}{\encodingdefault}{\sfdefault}{bx}{n}

\newcommand{\E}{\mathbb{E}}

\usepackage{bm}
\usepackage{amsfonts}
\usepackage{mathtools}
\usepackage{amsthm}
\usepackage{amsmath}
\usepackage{booktabs}
\usepackage{xcolor}         
\usepackage[T1]{fontenc}
\usepackage{arydshln}
\usepackage[hidelinks,colorlinks,citecolor=SpringGreen4]{hyperref}
\hypersetup{citecolor=[RGB]{0,139,69}}
\usepackage{enumitem}
\usepackage{url}
\usepackage{subcaption}
\usepackage{graphicx}
\usepackage{wrapfig}
\usepackage{multirow}
\usepackage{xspace}
\usepackage{interval}
\usepackage{colortbl}
\usepackage{pifont}
\usepackage{marvosym}

\newcommand{\ie}{\textit{i}.\textit{e}.}
\newcommand{\eg}{\textit{e}.\textit{g}.}

\newcommand{\reinforce}{\textsc{Reinforce}}
\definecolor{Gray1}{gray}{0.8}
\definecolor{Gray2}{gray}{0.9}

\newcommand{\paref}[1]{(\ref{#1})}

\title{Test-Time Adaptation with CLIP Reward for Zero-Shot Generalization in Vision-Language Models}

\author{Shuai Zhao\textsuperscript{\dag,\S}\thanks{Part of this work is done during an internship at Baidu Inc. Yi Yang is the corresponding author.} \quad Xiaohan Wang\textsuperscript{\P} \quad Linchao Zhu\textsuperscript{\ddag} \quad Yi Yang\textsuperscript{\ddag} \\
\textsuperscript{\dag}ReLER Lab, AAII, University of Technology Sydney \\
\textsuperscript{\ddag}ReLER Lab, CCAI, Zhejiang University \quad
\textsuperscript{\P}Stanford University \quad \textsuperscript{\S}Baidu Inc.\\
{\footnotesize
\texttt{\{zhaoshuaimcc, wxh1996111\}@gmail.com} \quad 
\texttt{\{zhulinchao, yangyics\}@zju.edu.cn}
}
}

\iclrfinalcopy 
\begin{document}

\maketitle

\begin{abstract}
One fascinating aspect of pre-trained vision-language models~(VLMs) learning under language supervision is their impressive zero-shot generalization capability.
However, this ability is hindered by distribution shifts between the training and testing data.
Previous test time adaptation~(TTA) methods for VLMs in zero-shot classification rely on minimizing the entropy of model outputs, tending to be stuck in incorrect model predictions.
In this work, we propose TTA with feedback to rectify the model output and prevent the model from becoming blindly confident.
Specifically, a CLIP model is adopted as the reward model during TTA and provides feedback for the VLM.
Given a single test sample,
the VLM is forced to maximize the CLIP reward between the input and sampled results from the VLM output distribution.
The proposed \textit{reinforcement learning with CLIP feedback~(RLCF)} framework is highly flexible and universal.
Beyond the classification task, with task-specific sampling strategies and a proper reward baseline choice, RLCF can be easily extended to not only discrimination tasks like retrieval but also generalization tasks like image captioning,
improving the zero-shot generalization capacity of VLMs.
According to the characteristics of these VL tasks, we build different fully TTA pipelines with RLCF to improve the zero-shot generalization ability of various VLMs.
Extensive experiments along with promising
empirical results demonstrate the effectiveness of RLCF.
The code is available at \url{https://github.com/mzhaoshuai/RLCF}.
\end{abstract}

\section{Introduction}
Pre-trained vision-language models~(VLMs) learning under language supervision~\citep{radford2021clip,jia2021align,yuan2021florence} exhibit promising zero-shot transferability.
This encourages researchers to explore the capabilities of VLMs across a number of tasks in a zero-shot fashion.
For example, \cite{HongZPCYL22} employ CLIP for zero-shot text-driven avatar generation,
\cite{DBLP:conf/cvpr/SainBCKXS23} adapt CLIP for zero-shot sketch-based image retrieval,
and \cite{li2023decap} achieve zero-shot image captioning without images.
Nonetheless, the large domain gap between training and test data is still challenging for VLMs in a zero-shot circumstance.
In this work, we investigate how to fulfill the domain gap during test time in various tasks without task-specific training corpus,
namely, test time adaptation~(TTA) for VLMs with a zero-shot prerequisite.

\begin{figure*}[!t]
\vspace{-0.5cm}
\centering
\begin{subfigure}{1.0\textwidth}
\centering
\includegraphics[width=1.0\linewidth]{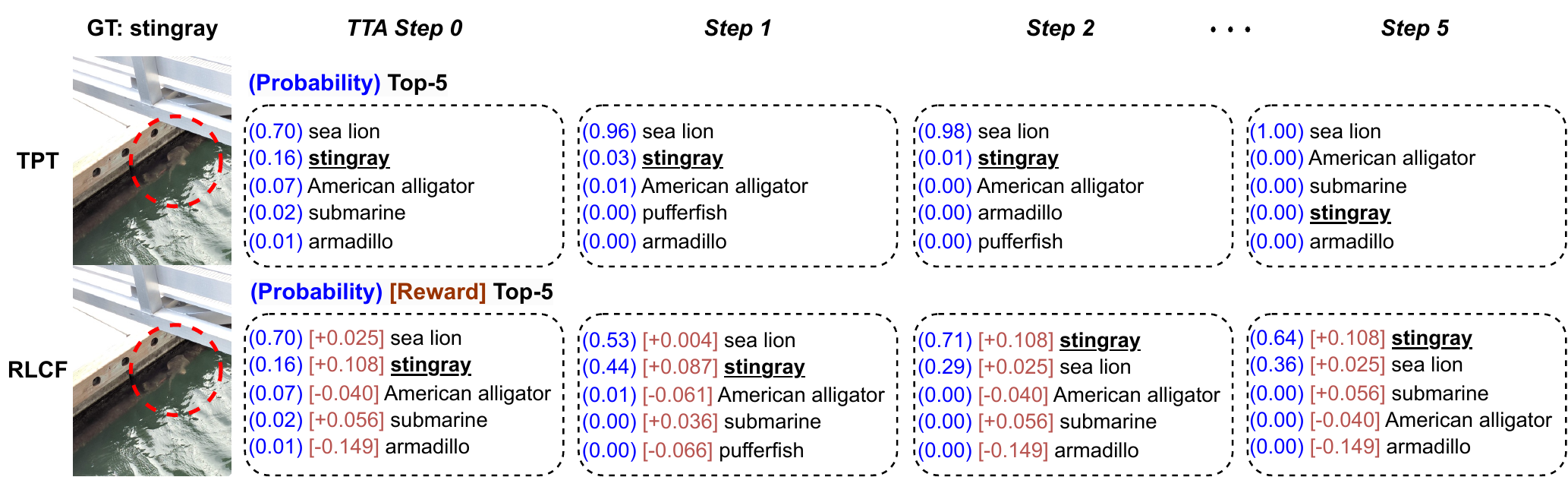}
    \caption{\textbf{Top-5 predictions and probabilities with different TTA steps}. Step 0 shows original predictions.
    }
    \label{fig1:top-5-tpt-rlcf}
\end{subfigure}
\begin{subfigure}{1.0\textwidth}
    \centering
    \includegraphics[width=1.0\linewidth]{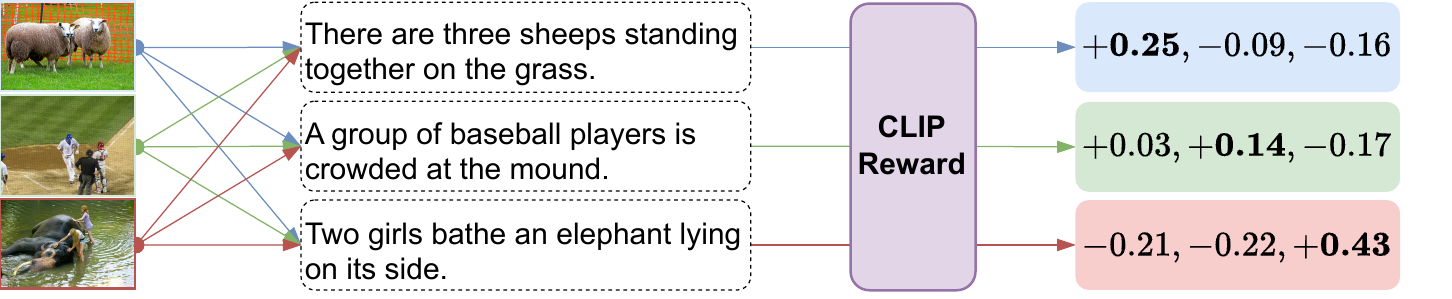}
    \caption{\textbf{Examples of CLIP reward}. The average score of an image and all sentences is the reward baseline.}
    \label{fig1:clip-reward}
\end{subfigure}
\caption{\textbf{Feedback mechanism in zero-shot generalization with CLIP as the reward model}.}
\label{fig1:diff-tpt-rlcf}
\vspace{-0.2cm}
\end{figure*}

One pioneer TTA work in improving the zero-shot classification ability of VLMs is test time prompt tuning~(TPT)~\citep{shu2022tpt}.
Given a single test sample, TPT optimizes the learnable prefix tokens by minimizing the entropy of model outputs to bootstrap its generalization capacity. Nevertheless, making the model confident in its predictions is a double-edged sword.
It does reduce the test error and close the domain gap at a certain level~\citep{wang2021tent}, but it makes the model stick to its incorrect predictions and unable to get out of the dilemma by itself as shown in the top of Figure~\ref{fig1:top-5-tpt-rlcf}.
Entropy minimization tends to make the model blindly confident.

Inspired by the success of the feedback mechanism in language and vision tasks~\citep{ouyang2022training,openai2023gpt,pinto2023tuning}, we introduce feedback during test time to rectify the VLM output as shown in the bottom of Figure~\ref{fig1:top-5-tpt-rlcf}.
Previous feedback methods leverage labeled preference data to train a reward model~\citep{ouyang2022training,lee2023rlaif}
or use labels to calculate the reward~\citep{Cho2022CLIPReward,pinto2023tuning}.
Without ground truth, we refer to the well-recognized CLIP~\citep{radford2021clip} model as the feedback resource.
CLIP shows powerful generalization capacity across many VL tasks.
The outputs of CLIP are also well-calibrated~(without fine-tuning on a specific dataset)~\citep{DBLP:conf/nips/MindererDRHZHTL21}, \ie, the score from CLIP accurately reflects its uncertainty about the input sample.
This makes CLIP a reliable reward model.
One more question is why feedback rather than directly tuning with CLIP supervision? \cite{ouyang2022training} demonstrate that model learning with feedback has better generalization abilities than a supervised fine-tuning model.
We get the same conclusion from our empirical results.
Furthermore, CLIP supervision cannot be directly used in generation tasks like image captioning, while the feedback mechanism is versatile.

Our proposed framework, coined as \textit{reinforcement learning with CLIP feedback~(RLCF)}, is flexible and universal for TTA with different VLMs in various tasks.
With task-specific sampling strategies and a proper reward baseline choice, RLCF is applicable across zero-shot classification, text-image retrieval, and image captioning.
In these tasks, the model is given a single test sample, we then sample $K$ candidates from the output distribution.
For discrimination tasks like classification and retrieval, the top-$K$ sampling is applied; for the caption generalization task, a beam search method is adopted.
Assuming the input is an image, like Figure~\ref{fig1:clip-reward}, the CLIP model first gives the CLIPScore~\citep{hessel2021clipscore} between the image and all candidate sentences.
As CLIPScore is always non-negative, the average score is subtracted from the calculated scores.
This average baseline aims to distinguish which model behaviors are encouraged and which are discouraged.
Then the learnable parameters in the TTA model are optimized by \reinforce{}~\citep{williams1992reinforce} algorithm.

While the reward design and learning algorithm remain consistent across various tasks, the TTA pipelines are tailored to each specific task.
For classification, we inherit the data augmentation and confidence selection pipeline from TPT~\citep{shu2022tpt}, making it work for not only prefix tuning but also backbone adaptation.
For retrieval, considering a large number of candidate entries, we only update the parameters \textit{w.r.t.} the query for efficiency.
For instance, we only tune the branch \textit{w.r.t.} the input modality for a two-branch VLM like CLIP.
For image captioning, we construct the TTA pipeline with two methods~\citep{mokady2021clipcap,nukrai2022text} built upon large language models~(LLMs).
During TTA, we only tune the projector which projects the image into the LLM token embedding space.
Plus, several task-agnostic practical tricks are applied, \ie, multiple reward models, episodic TTA~\citep{wang2021tent}, and momentum buffer for incremental learning.

To summarize our contributions:
1) To the best of our knowledge, RLCF is the first universal fully TTA framework for improving the zero-shot generalization capacity of VLMs across different tasks.
2) We develop a novel reward function for test time RL with CLIP. It is simple yet effective.
Compared to previous methods~\citep{Cho2022CLIPReward} in the training stage, it demonstrates that CLIP can be used as a practical reward model alone, even with a single test sample.
3) We design task-specific TTA pipelines for three VL tasks with RLCF.
Extensive experiments with promising results validate the effectiveness of RLCF in boosting the zero-shot performance of different VLMs.

\section{Related Work}
\label{sec:relatedwork}

\textbf{Reinforcement learning in language and vision~~}
The most well-known application of RL in natural language process is reinforcement learning with human feedback~(RLHF)~\citep{ouyang2022training,openai2023gpt}.
A reward model is trained with preference data collected from humans, and it is used to fine-tune the LLM via proximal policy optimization~(PPO,~\cite{schulman2017proximal}).
Similar approaches are applied in~\citep{ziegler2019fine,stiennon2020learning,bai2022training,glaese2022improving}.
In prompt engineering for language models,
RLPrompt~\citep{deng-etal-2022-rlprompt}
and TEMPERA~\citep{zhang2023tempera}
search for discrete text prompts by RL.
RL has also been widely applied in vision and multi-modal research.
A comprehensive study of deep RL in computer vision can be found at~\citep{le2022deep}.
Recently, Pinto~\textit{et al.}~\citep{pinto2023tuning} optimize vision task metrics using RL and achieve promising results, demonstrating the effectiveness of RL in vision.
In the multi-modal area, 
ImageReward~\citep{xu2023imagereward}
collects preference data and trains a reward model for text-to-image generation tasks, similar to RLHF in GPT models.
SCST~\citep{rennie2017self} apply CIDEr
metric as a reward function with \reinforce{}~\citep{williams1992reinforce} algorithm to improve the generation quality in image captioning during training.
Cho~\textit{et al.}~\citep{Cho2022CLIPReward} explore the possibility of using CLIPScore~\citep{hessel2021clipscore}
as the reward in image captioning.
Their empirical results show that CLIPScore
cannot be an independent reward function and should be combined with a grammar regularization or CIDEr metric.
Nevertheless, the training of the grammar head and calculation of CIDEr metric both need reference text, which is unavailable at test time.

\textbf{Test-time adaptation~~}
Test-time adaptation~(TTA) aims to address the distribution shift
between training and test data during test time~\citep{sun2020ttt,liu2021ttt++,wang2021tent}.
Test-time training~(TTT, ~\cite{sun2020ttt})
allows modifications to the training pipeline.
In such cases, self-supervised auxiliary tasks are incorporated to help the model adapt to the distribution of test data~\citep{sun2020ttt,liu2021ttt++,lin2022video}.
For example, TTT+~\citep{liu2021ttt++} utilizes instance discrimination tasks from contrastive learning~\citep{chen2020simple}.
On the other hand, fully TTA assumes that the training pipeline cannot be modified as the training data is unavailable~\citep{wang2021tent}.
Two popular techniques in fully TTA are normalization layer adaptation and entropy minimization.
Normalization layer adaptation updates data statistics or parameters of the normalization layer based on batched test samples~\citep{wang2021tent,schneider2020improving,niu2023towards} or augmented data views from a single test sample~\citep{zhang2022memo}.
Entropy minimization aims to make the model confident in its predictions to reduce generalization error~\citep{wang2021tent,zhang2022memo,shu2022tpt,niu2022efficient,niu2023towards}.
There is also a retrieval-augmented TTA method~\citep{zancato2023train}, which uses CLIP to retrieve informative data from an external large dataset and update the decision boundary during test time.

\section{Method}

\subsection{Preliminaries}
\textbf{Fully test-time adaptation in vision-language tasks~~}
Let $f_{\theta}(\cdot)$ represent a VLM trained
on image-text pairs
$\mathcal{D}_{train} = \{(\bm{t}_i, \bm{v}_i)\}_{i=1}^N$ with
parameter $\theta$, where $\bm{t}_i \in \mathcal{T}_{train}$
(training text space) and $\bm{v}_i \in \mathcal{V}_{train}$ (training image space).
The objective of TTA~\citep{sun2020ttt,wang2021tent} is to boost $f_{\theta}(\bm{t})$ or
$f_{\theta}(\bm{v})$ on domain-shifted test samples
$\mathcal{D}_{test} = \{(\bm{t}_j\}_{j=1}^M$
or 
$\mathcal{D}_{test} = \{(\bm{v}_j\}_{j=1}^M$,
where $\bm{t}_j \in \mathcal{T}_{test}$~(testing text space),
$\bm{v}_j \in \mathcal{V}_{test}$
(testing image space),
$\mathcal{T}_{test} \neq \mathcal{T}_{train},$
and
$\mathcal{V}_{test} \neq \mathcal{V}_{train}$.
We assume that the VLM
takes either text or image as input and outputs the other modality.
In fully TTA, the training data are unavailable, and the training pipeline cannot be modified.
Following TPT~\citep{shu2022tpt} and MEMO~\citep{zhang2022memo}, the adaptation is conducted with a single test point, \ie, the VLM is exposed to only one $\bm{t}_j$ or $\bm{v}_j$.

\textbf{Contrastive Language-Image Pre-training (CLIP)~~}
CLIP~\citep{radford2021clip} comprises an image encoder
${g}(\cdot)$ and a text encoder ${h}(\cdot)$.
CLIP is pre-trained using a contrastive loss that
encourages similarity between feature vectors of paired images and text, aligning them in a shared embedding space.
Once pre-trained, CLIP can assess the similarity between the text $\bm{t}$ and image $\bm{v}$ as follows:
\begin{align}
  \texttt{CLIP}(\bm{t}, \bm{v}) = \cos(h(\bm{t}), g(\bm{v})),
\end{align}
where $\cos(\cdot,\cdot)$ represents the cosine similarity.
For image classification with CLIP, the input text consists of the prompt plus the class names, \ie,
$\bm{t} = \{\bm{p}_t; \text{"dog"}\}$, where
prompt $\bm{p}_t = \text{"a photo of a"}$.

\subsection{Test-time adaptation with CLIP reward}
\subsubsection{Reinforcement learning with CLIP feedback}
Without loss of generality, we first consider the case where the VLM $f_{\theta}(\cdot)$ takes an image $\bm{v}$ as input and maps it to text $\bm{t}$.
During TTA, our goal is to learn a conditional distribution $P(\bm{t}|\bm{v}, \theta)=f_{\theta}(\bm{v})$ that maximizes a reward function
$\mathcal{R}(\cdot,\cdot)$.
Formally, the optimization problem during TTA is:
\begin{align} \label{optim_goal}
\max\limits_\theta \E_{\bm{t} \sim P(\cdot|\bm{v},\theta)} \mathcal{R}(\bm{t}, \bm{v}).
\end{align}
Different from previous methods~\citep{rennie2017self,Cho2022CLIPReward,pinto2023tuning}
which maximizes the expected reward over batched training
samples, here we only maximize the expected reward over \textit{a single test sample} $\bm{v} \in \mathcal{V}_{test}$.

\textbf{Policy gradient with \reinforce{}~~}
To compute the gradient of the non-differentiable reward function, \reinforce{}~\citep{williams1992reinforce} is adopted to calculate $\nabla_\theta \E_{\bm{t}\sim P}[\mathcal{R}(\bm{t}, \bm{v})]$.
It uses the so-called "log-derivative trick" to estimate
the gradient of the expected reward for a given input:
\begin{align} \label{eq:reinforce}
\nabla_\theta \E_{\bm{t} \sim P} [ \mathcal{R}(\bm{t}, \bm{v}) ] 
= \E_{\bm{t} \sim P} [ \mathcal{R}(\bm{t}, \bm{v}) \nabla_\theta \log P(\bm{t}|\bm{v}; \theta) ].
\end{align}
In a VL task,
the input and output modalities are closely related, \eg, the input is an image and the output is the description of the image.
Therefore, we can use CLIP to evaluate the
similarity between the input and output, and the model can maximize this similarity to align with task goals. Similar to~\cite{Cho2022CLIPReward}, we use
CLIPScore~\citep{hessel2021clipscore} as the reward:
\begin{align}
\texttt{CLIP-S}(\bm{t}, \bm{v}) = w\times\max(\texttt{CLIP}(\bm{t}, \bm{v}), 0)
,
\end{align}
where $w=2.5$ is a constant.
CLIPScore is always non-negative, which means it encourages all model behaviors.
However, for an irrelevant sampled image-text pair in
Figure~\ref{fig1:clip-reward}, we expect the reward model
to provide negative feedback to discourage such behavior.
Cho~\textit{et al.}~\citep{Cho2022CLIPReward} adopt a greedy search baseline which needs to be combined with a grammar regularization or CIDEr metric to be a practical reward function.
In this work, we demonstrate that with proper sampling
strategies and baseline, CLIPScore can also be used as the sole reward function in different VL tasks.
Specifically, we set the reward baseline as the average
CLIPScore of sampled image-text pairs. The reward
function with baseline becomes:
\begin{align} \label{eq:reward}
\mathcal{R}(\bm{t}, \bm{v})
= \texttt{CLIP-S}(\bm{t}, \bm{v}) - \E_{\bm{t} \sim P} [ \texttt{CLIP-S}(\bm{t}, \bm{v}) ].
\end{align}
It is straightforward to get the reward function for a VLM which takes text $\bm{t}$
as input and return an image $\bm{v}$ according to Eq.~\paref{eq:reward}.
The sampling strategies will be presented in the next section.

\begin{figure}[!t]
\vspace{-0.5cm}
\centering
\resizebox{0.98\linewidth}{!}{
\includegraphics{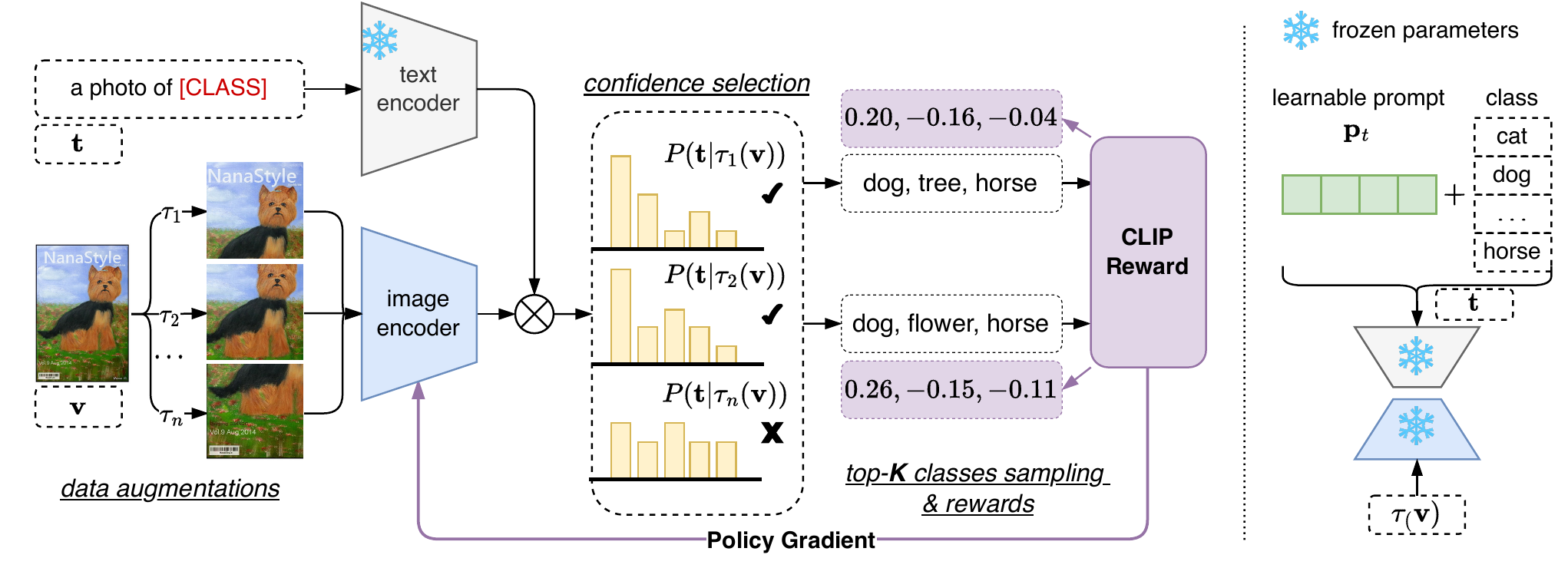}
}
\caption{
\textbf{Fully TTA for zero-shot image classification with CLIP reward.} \textit{Left}: image encoder tuning.
\textit{Right}: prompt tuning.
The pipelines of the two are the same except
for the learnable parameters.
A single test image is first augmented to
produce multiple views,
then only confident views
with low-entropy predictions are selected.
For each selected view, we sample the top-$K$ classes, calculate their rewards, and update
the parameters using policy gradient.
}
\label{fig:tta-cls-ood}
\vspace{-0.2cm}
\end{figure}

\begin{figure}[!t]
\centering
\resizebox{0.98\linewidth}{!}{
\includegraphics{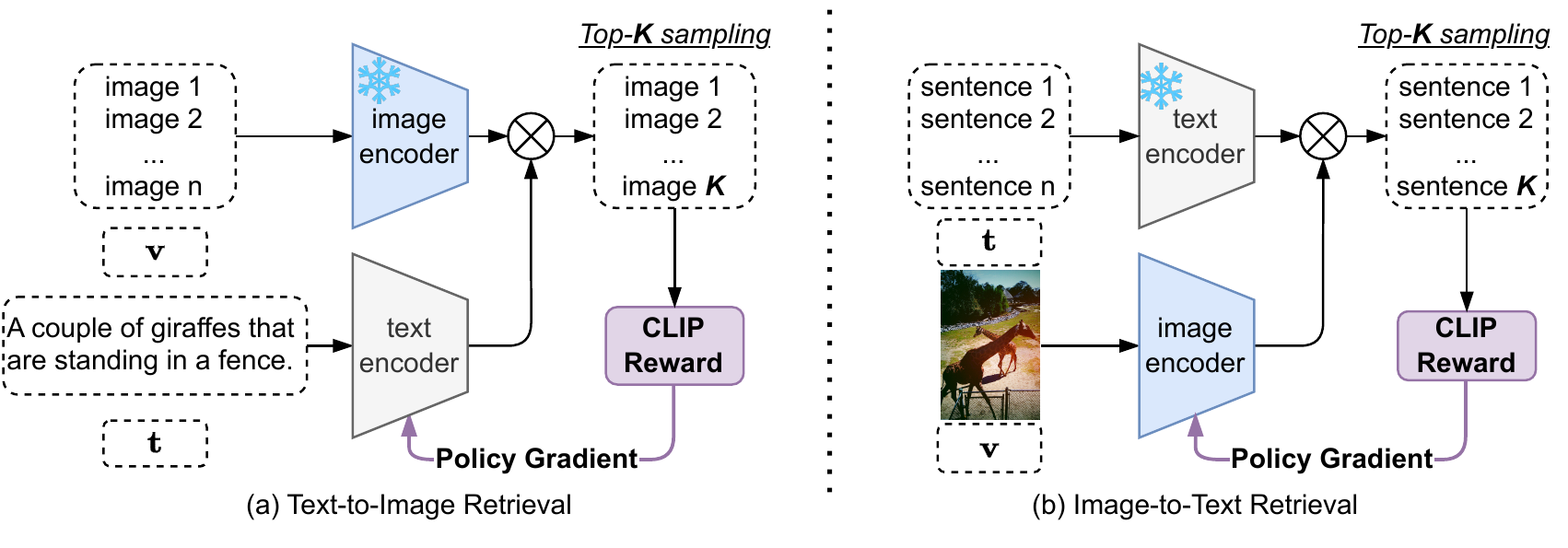}
}
\caption{
\textbf{Fully TTA for zero-shot text-image retrieval with CLIP reward}.
}
\label{fig:tta-ret}
\vspace{-0.3cm}
\end{figure}

\subsubsection{Task-specific fully test-time adaptation}
\label{sec:task-tta}
RLCF is flexible and applicable across various VL tasks, and we apply RLCF to three different VL tasks in this work.
For all tasks, the VLM solely learns through
\reinforce{} with Eq.~\paref{eq:reward} as the reward function during test time.
However, VLMs and sampling strategies vary with tasks.
Next, we introduce our task-specific fully TTA pipelines.

\textbf{Zero-shot image classification on OOD data~~}
Figure~\ref{fig:tta-cls-ood} illustrates the fully TTA pipeline for zero-shot image classification with RLCF.
Without loss of generality, we also choose CLIP as the classifier.
The TTA pipelines include two adaptation manners: prompt tuning and image encoder tuning.
TPT~\citep{shu2022tpt} shows that entropy minimization for image encoder tuning results in inferior performance compared to prompt tuning.
By contrast, RLCF works both with prompt tuning and image encoder tuning, demonstrating its versatility.

In Figure~\ref{fig:tta-cls-ood},
given a test image $\bm{v}$, it is first
operated with data augmentors $\{\tau_1, \tau_2, \ldots, \tau_n\}$ for multiple different views.
Following TPT~\citep{shu2022tpt}
and SAR~\citep{niu2023towards}, we only reserve the
confident samples with low-entropy predictions,
namely, the entropy $H(P(\bm{t}|\tau(\bm{v})))$ of
the selected view should be low.
High-entropy predictions are considered unreliable as they lack confidence in their outputs.
In practice, we use the bottom $10$th percentile 
of $n=64$ augmented views with low entropies as TPT~\citep{shu2022tpt}.
For each low-entropy view, class names of the top-$K$ predictions are used to calculate their CLIP rewards
according to Eq.~\paref{eq:reward}.
The learnable parameters are then optimized to maximize
the expected reward by gradient descent as Eq.~\paref{eq:reinforce}.

One point that needs to be clarified is why using the CLIP reward as feedback rather than directly fine-tuning the model with CLIP supervision.
For example, methods like knowledge distillation~(KD,~\cite{hinton2015distilling}) or pseudo-label~\citep{lee2013pseudo}.
InstructGPT~\citep{ouyang2022training} demonstrates that model learning with feedback has better generalization capabilities compared to a supervised fine-tuning model.
In our context, KD or pseudo-label requires a weak model~(student) to mimic a strong model~(teacher).
However, it is worth noting that the student may be correct while the teacher may be incorrect.
For instance, given an image of a dog, 
the top-3 predictions of the student and teacher models are \{dog, horse, tree\} and \{cat, dog, horse\}, respectively.
For KD or pseudo-label, the student will be forced to follow the incorrect behaviors of the teacher.
In contrast, the feedback mechanism only assesses the sampled results from the student, less likely to alter the correct prediction.
In such cases, the feedback mechanism combines the merits of both the student and the teacher.
Another important reason is that CLIP supervision cannot be directly used in generalization tasks like image captioning, while the feedback mechanism is universal.

\textbf{Zero-shot text-image retrieval~~}
The fully TTA pipeline for zero-shot retrieval with RLCF is presented in Figure~\ref{fig:tta-ret}.
CLIP also serves as the zero-shot retrieval model.
For retrieval, the number of candidates is usually large, so we only update the parameters with respect to the query for efficiency.
For text-to-image retrieval, the image encoder remains fixed, while the text encoder is frozen in the other case.
Given a query, top-$K$ sampling is employed to the returned results to calculate the reward.
Unlike image classification, no augmentations are used for the input query.
The retrieval task requires a holistic understanding of the input query rather than identifying a single object. Augmentations like crop and flip may lead to corrupt semantics.

\begin{figure}[!t]
\vspace{-0.5cm}
\centering
\resizebox{1.0\linewidth}{!}{
\includegraphics{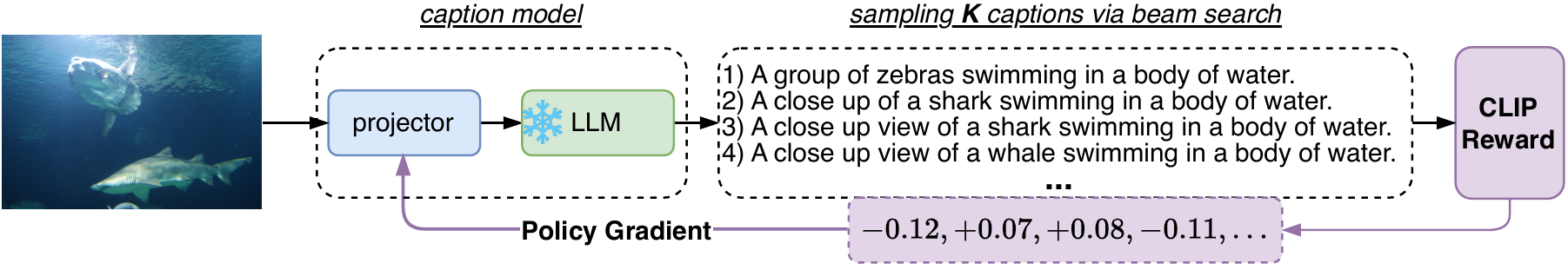}
}
\caption{
\textbf{Fully TTA for image captioning with CLIP reward.} 
}
\label{fig:tta-cap}
\vspace{-0.3cm}
\end{figure}

\textbf{Zero-shot and cross-domain image captioning~~}
Figure~\ref{fig:tta-cap} illustrates the fully TTA pipeline for image captioning with RLCF.
The captioning TTA pipeline is built upon two LLM-based methods: CapDec~\citep{nukrai2022text} and CLIPCap~\citep{mokady2021clipcap}.
CapDec is trained only with text and CLIPCap is trained with images.
TTA with CapDec is undertaken with a zero-shot prerequisite, and TTA with CLIPCap is cross-domain.
During the test, CapDec and CLIPCap will be given unseen and domain-shifted images, respectively.
Both CapDec and CLIPCap utilize a projector~(\eg, an MLP or
transformer~\citep{vaswani2017attention}) to project
CLIP feature vectors into the token embedding space of the LLM.
Only the projector is updated through policy gradient, while the LLM remains fixed during TTA.
Beam search is employed to sample $K$ generated captions for reward calculation.

\subsubsection{Test-time adaptation tricks}
In this section, we introduce several general TTA techniques applicable across different tasks.

\textbf{Multiple reward models with weights~~}
By default, a single CLIP-ViT-L/14 is used as the reward model.
An ensemble of multiple reward models can be used for better feedback.
We assign scores based on human preference for different CLIP models:
\{CLIP-ViT-L/14-336: 10, CLIP-ViT-L/14: 5, CLIP-RN50$\times$64: 3\}.
These scores are then normalized to sum up to 1, serving as weights for the ensemble.
CLIP-RN uses a ResNet~\citep{resnet} as the image encoder, while CLIP-ViT adopts a vison transformer~\citep{dosovitskiy2021an}.

\textbf{Episodic TTA~~} The model is exposed to only a single test sample once,   making the learned knowledge unreliable for other samples.
Hence, after each TTA process, the model parameters $\theta$ are reset to the initial state $\theta^\star$ like~\citep{wang2021tent,shu2022tpt}.
It is called episodic TTA.
  
\textbf{Momentum buffer~~}
While episodic TTA ensures reliability, it limits the incremental learning ability of the model.
To address this issue, we introduce a momentum buffer
$\xi$, initialized as $\xi \leftarrow \theta^\star$.
  After a TTA process, $\theta$ becomes $\overline{\theta}$, and
  $\xi$ is updated by
  $\xi \leftarrow m \xi + (1-m)\overline{\theta}$, where
$m\in [0, 1)$ is a momemtum coefficient.
Every $B_s$ samples, we update $ \theta^\star \leftarrow \xi$.
At the start of the next TTA process, 
$\theta \leftarrow \theta^\star$, allowing the utilization of the learned knowledge.
The momentum buffer functions similarly to an ensemble of different models, resembling model soups~\citep{wortsman2022model}.

\section{Experiments} \label{sec:exp}

\begin{table}[!t]
\vspace{-0.5cm}
  \caption{\textbf{Top-1 accuracy of zero-shot image classification with TTA on OOD data}.
    KD uses a CLIP-ViT-L/14 as the teacher.
The\colorbox{Gray1}{best}and\colorbox{Gray2}{second-best}results are highlighted.
  Improvement in accuracy of RLCF compared to the baselines~(zero-shot CLIP-ViT-B/16 or CoOp) is in \textcolor{blue}{(${\uparrow}$blue)}.
  }
  \label{tab:ood-main}
  \centering
  \resizebox{0.96\linewidth}{!}{%
  \begin{tabular}{l|c|*{5}c}
    \toprule
    Method
    & ImageNet          &  ImageNet-A 
    & ImageNet-V2       & ImageNet-R
    & ImageNet-Sketch   & OOD Average \\
    \midrule

    & \multicolumn{6}{c}{\textit{\underline{Zero-shot baseline}}}  \\
    CLIP-ViT-B/16  
    &  66.73 &  47.87  &   60.86    
    &  73.98 &  46.09  &   57.20  \\
    CLIP-ViT-L/14  
    &  73.44 &  68.82  &   67.80    
    &  85.40 &  57.84  &   69.97  \\
  
    \midrule
    & \multicolumn{6}{c}{\underline{\textit{Prompt tuning for} CLIP-ViT-B/16}} \\
    CoOp~\citep{zhou2021coop}
    &  71.51  &  49.71  &   64.20
    &  75.21  &  47.99  &   59.28  \\


    CoCoOp~\citep{zhou2022cocoop}
    &  71.02  & 50.63  & 64.07      
    &  76.18   &  48.75  & 59.91    \\
    
    TPT~\citep{shu2022tpt} 
    &  68.98  & 54.77  & 63.45      
    &  77.06  & 47.94 &  60.81  \\

    TPT~+~CoOp~(\citeauthor{shu2022tpt})    
    & 73.61  & 57.95  & 66.83
    & 77.27  & 49.29  & 62.84 \\

    TPT~+~CoOp~+~KD~(\citeauthor{hinton2015distilling})
    &  71.40  &  63.25  &   65.28    
    &  82.70  &  55.78  &   66.75 \\

    \textbf{RLCF}
    &73.23\textcolor{blue}{$_{(\uparrow6.50)}$} 
    &65.45\textcolor{blue}{$_{(\uparrow17.58)}$}  
    &69.77\textcolor{blue}{$_{(\uparrow8.91)}$} 
    &83.35\textcolor{blue}{$_{(\uparrow9.37)}$}  
    &54.74\textcolor{blue}{$_{(\uparrow8.65)}$}  
    &68.33\textcolor{blue}{$_{(\uparrow11.13)}$}  \\

    \textbf{RLCF + CoOp}
    & \cellcolor{Gray2}76.05\textcolor{blue}{$_{(\uparrow4.54)}$}
    &  \cellcolor{Gray2}69.74\textcolor{blue}{$_{(\uparrow20.03)}$}
    &  \cellcolor{Gray2}70.62\textcolor{blue}{$_{(\uparrow6.42)}$}
    &  \cellcolor{Gray2}84.51\textcolor{blue}{$_{(\uparrow9.30)}$}
    &  \cellcolor{Gray2}56.49\textcolor{blue}{$_{(\uparrow8.50)}$}
    &  \cellcolor{Gray2}70.34\textcolor{blue}{$_{(\uparrow11.06)}$}  \\

    \textbf{RLCF-S + CoOp}
    &  \cellcolor{Gray1}76.50\textcolor{blue}{$_{(\uparrow4.99)}$}
    & \cellcolor{Gray1}71.11\textcolor{blue}{$_{(\uparrow21.40)}$}
    &   \cellcolor{Gray1}70.92\textcolor{blue}{$_{(\uparrow6.72)}$}
    &  \cellcolor{Gray1}84.73\textcolor{blue}{$_{(\uparrow9.52)}$}
    &  \cellcolor{Gray1}56.97\textcolor{blue}{$_{(\uparrow8.98)}$}
    &\cellcolor{Gray1}70.93\textcolor{blue}{$_{(\uparrow11.65)}$}  \\

    \midrule
    & \multicolumn{6}{c}{\underline{\textit{Image encoder tuning for} CLIP-ViT-B/16}} \\

    Pseudo-label~\citep{lee2013pseudo}
    &  69.11  &  62.15  &   63.56    
    &  80.03  &  49.45  &   63.80  \\

    TPT~\citep{shu2022tpt}
    &  69.42  &  61.62  &   63.70    
    &  79.74  &  49.47  &   63.63  \\
    
    KD~\citep{hinton2015distilling}
    &  70.92  &  66.39  &   65.01    
    &  82.12  &  53.51  &   66.76  \\

    ATKD~\citep{guo2020reducing}
    &  70.51  &  70.66  &   65.54    
    &  85.12  &  53.56  &   68.72  \\

    \textbf{RLCF}
    &  74.85\textcolor{blue}{$_{(\uparrow8.12)}$} 
    &  73.71\textcolor{blue}{$_{(\uparrow25.84)}$}  
    &  69.77\textcolor{blue}{$_{(\uparrow8.91)}$} 
    &  86.19\textcolor{blue}{$_{(\uparrow12.21)}$}  
    &  57.10\textcolor{blue}{$_{(\uparrow11.01)}$}  
    & 71.69\textcolor{blue}{$_{(\uparrow14.49)}$}  \\

    \textbf{RLCF-S}
    & \cellcolor{Gray2}75.34\textcolor{blue}{$_{(\uparrow8.61)}$}
    & \cellcolor{Gray2}75.00\textcolor{blue}{$_{(\uparrow27.13)}$}  
    & \cellcolor{Gray2}70.08\textcolor{blue}{$_{(\uparrow9.22)}$}    
    & \cellcolor{Gray2}86.97\textcolor{blue}{$_{(\uparrow12.99)}$}  
    & \cellcolor{Gray1}57.75\textcolor{blue}{$_{(\uparrow11.66)}$} 
    & \cellcolor{Gray2}72.45\textcolor{blue}{$_{(\uparrow15.25)}$}  \\  

    \textbf{RLCF-S-M}
    & \cellcolor{Gray1}75.48\textcolor{blue}{$_{(\uparrow8.75)}$}  
    & \cellcolor{Gray1}75.16\textcolor{blue}{$_{(\uparrow27.29)}$} 
    & \cellcolor{Gray1}70.42\textcolor{blue}{$_{(\uparrow9.56)}$}    
    & \cellcolor{Gray1}87.23\textcolor{blue}{$_{(\uparrow13.25)}$}  
    & \cellcolor{Gray2}57.73\textcolor{blue}{$_{(\uparrow11.64)}$} 
    & \cellcolor{Gray1}72.64\textcolor{blue}{$_{(\uparrow15.44)}$}   \\  
    \bottomrule
  \end{tabular}
}
\vspace{-0.3cm}
\end{table}

  

This section presents the experimental TTA results
in three tasks.
For variants of our method,
\textbf{RLCF} uses a CLIP-ViT-L/14 as the reward model,
\textbf{RLCF-S} adopts weighted reward sum of \{CLIP-ViT-L/14-336, CLIP-ViT-L/14, CLIP-RN50$\times$64\},
and \textbf{RLCF-S-M} adds the momentum buffer.

\subsection{Zero-shot Image classification on OOD data}
\label{sec:exp-ood}
\textbf{Datasets~~}
Following CLIP and TPT, we test RLCF on ImageNet~\citep{imagenet_cvpr09} and its four 
variant test sets with distribution shifts:
ImageNet-A~\citep{Hendrycks2021ima},
ImageNet-V2~\citep{Recht2019imv2},
ImageNet-R~\citep{Hendrycks2021imr},
and
ImageNet-Sketch~\citep{wang2019learning}.
ImageNet-A consists 7,500 natural adversarial images misclassified by a ResNet-50.
ImageNet-V2 contains 10,000 natural images from different sources.
ImageNet-R collects 30,000 images with artistic renditions. ImageNet-Sketch includes 50,000 black and white sketch images.

\textbf{Baselines~~}
We compare RLCF with few-shot prompt tuning methods for CLIP --- CoOp~\citep{zhou2021coop} and
CoCoOp~\citep{zhou2022cocoop} (16 shots on ImageNet), state-of-the-art test-time prompt tuning methods --- TPT~\citep{shu2022tpt}, and knowledge distillation~(KD~\citep{hinton2015distilling}, {ATKD~\citep{guo2020reducing}}), which use the reward model as the
teacher during test time.
TPT + CoOp means TPT adopts the learned prompts of CoOp as the initialization, otherwise, TPT uses token embedding of a hard prompt "a photo of a" as initial weights. For all prompt tuning methods,
the length of learnable prompts is 4.
Results of Pseudo-label~\citep{lee2013pseudo} are also presented.

\begin{figure*}[!t]
\centering
\begin{subfigure}{0.328\textwidth}
\centering
\includegraphics[width=\linewidth]{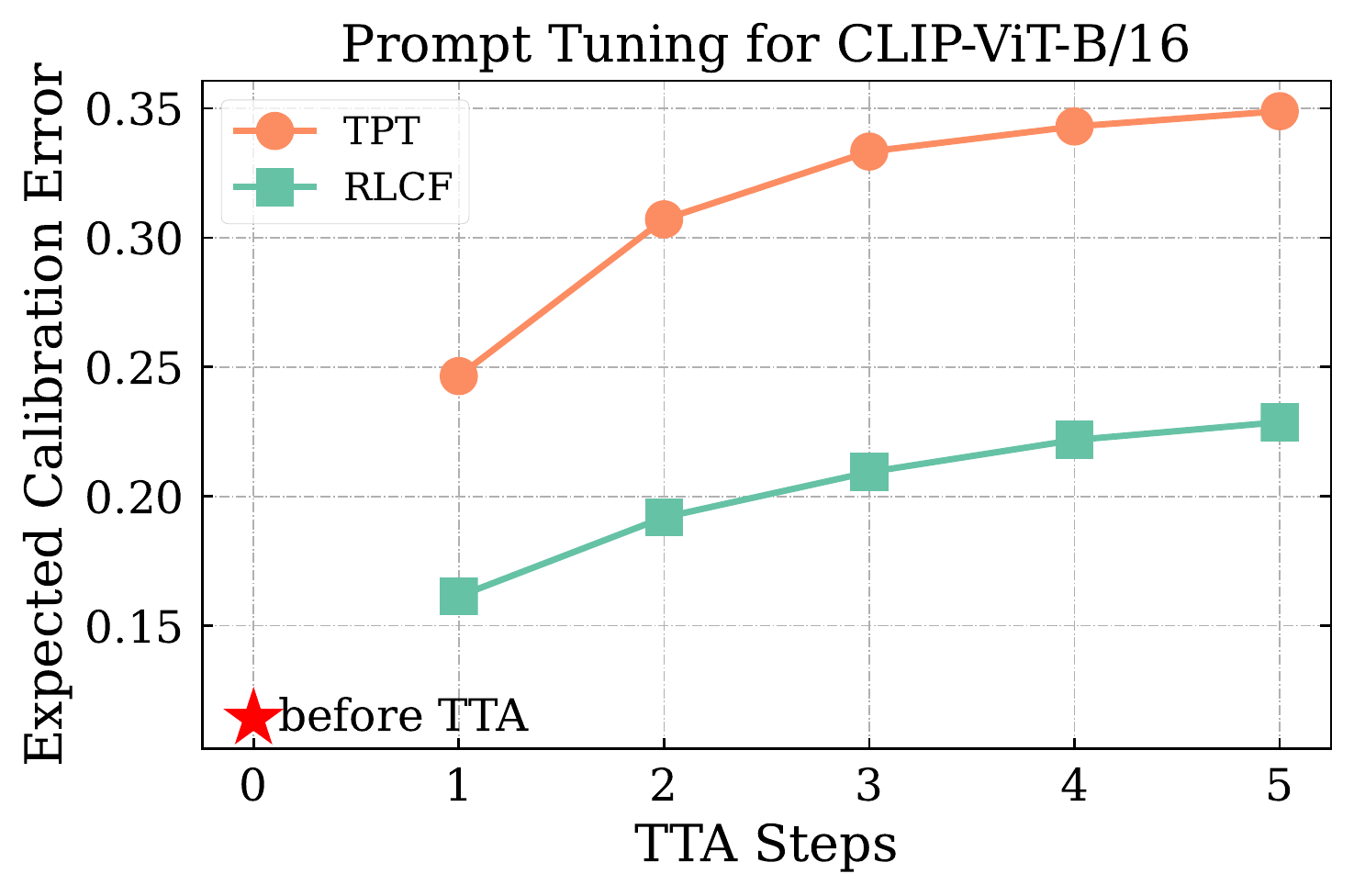}
  \caption{\small{ECE} on IN-A. lower is better.}
  \label{fig-ece}
\end{subfigure}
\begin{subfigure}{0.328\textwidth}
\centering
\includegraphics[width=\linewidth]{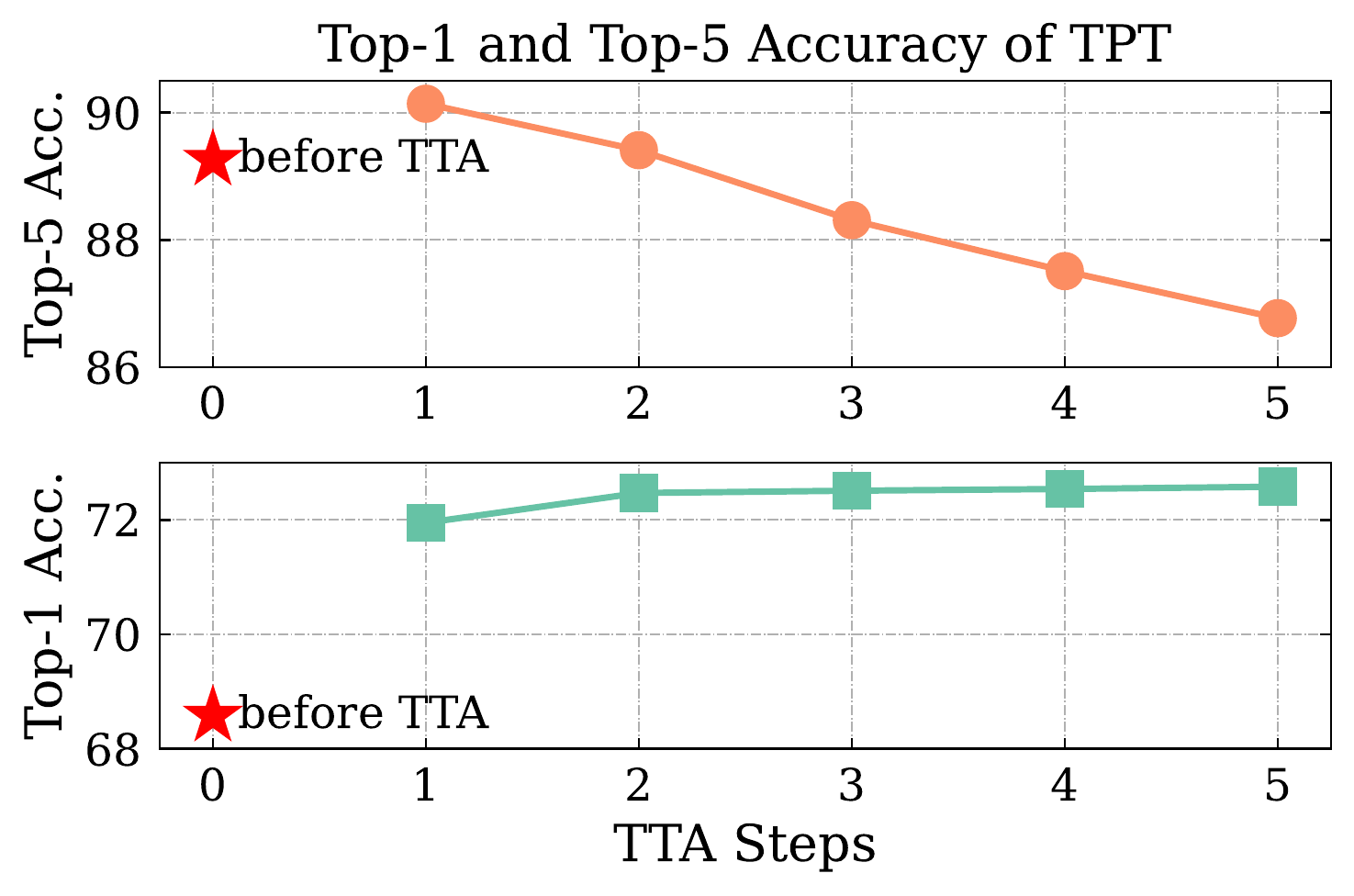}
\caption{\small{Acc. of TPT \textit{w.r.t.} steps.}}
\label{fig-tpt-acc-steps}
\end{subfigure}
\begin{subfigure}{0.328\textwidth}
\centering
\includegraphics[width=\linewidth]{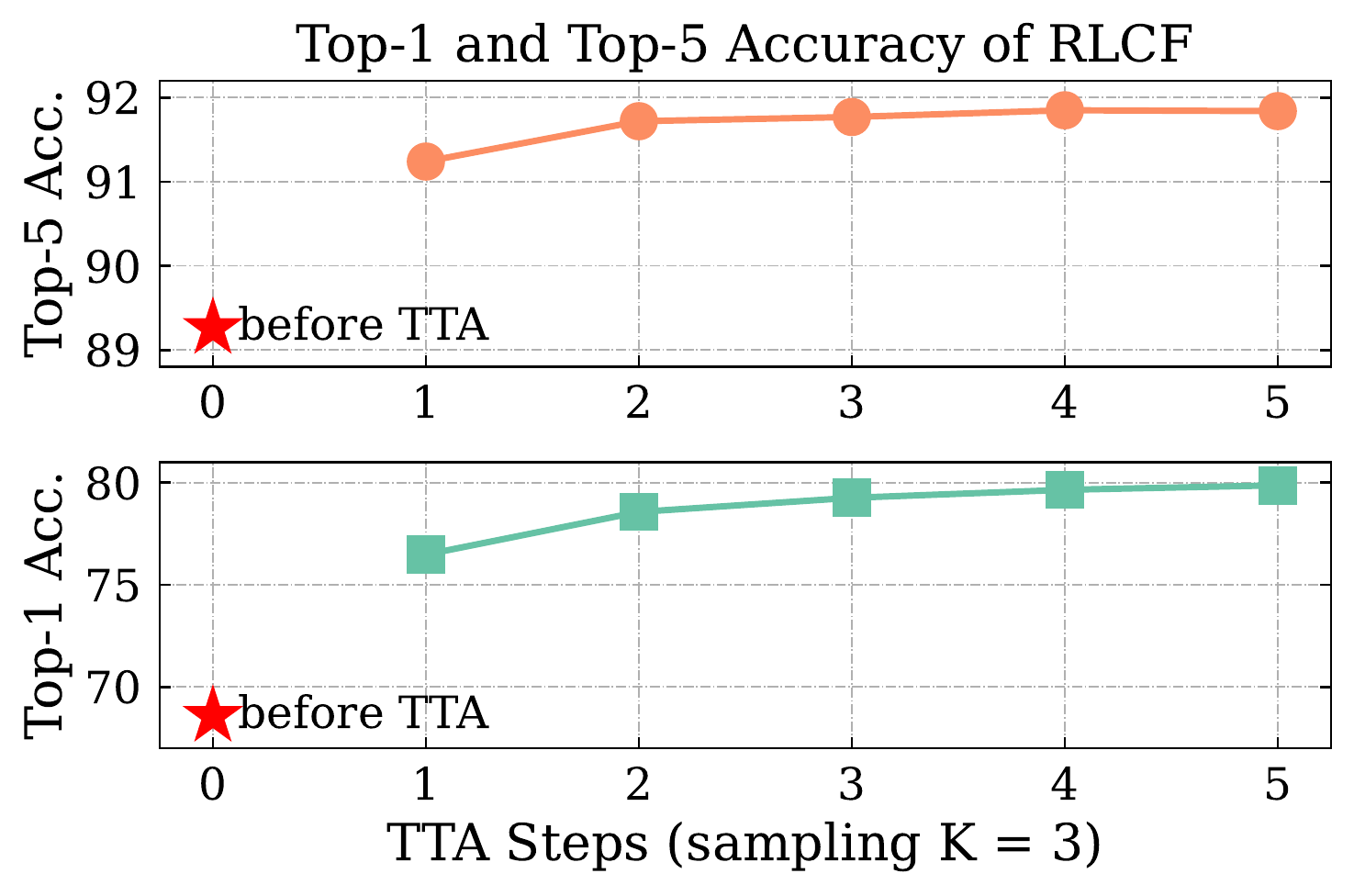}
\caption{\small{Acc. of RLCF \textit{w.r.t.} steps.}}
\label{fig-rlcf-acc-steps}
\end{subfigure}
\caption{\textbf{ECE and average accuracy on ImageNet-A/V2/R.} Prompt tuning with CLIP-ViT-B/16.
}
\label{fig:steps-k}
\vspace{-0.3cm}
\end{figure*}


\textbf{Implementation details~~}
For prompt tuning, the learning rate is 7e-3, the weight decay value is 5e-4, and the optimizer is AdamW~\citep{loshchilov2018decoupled}.
For image encoder tuning, the learning rate is decreased to 1e-5.
Given a test sample, the parameters will be optimized for 3 steps to maximize the reward of the top-$3$
(sampling factor $K=3$) predictions.
The momentum coefficient $m=0.9998$ and update interval
$B_s=64$ for the momentum buffer.

\textbf{Results~~}
In Table~\ref{tab:ood-main}, RLCF largely improves the zero-shot generalization capacity of CLIP-ViT-B/16 and outperforms previous methods.
Notably, on ImageNet-A/V2/R, \textit{RLCF with CLIP-ViT-B/16 surpasses the reward model --- CLIP-ViT-L/14}.
This shows that RLCF effectively combines the capabilities of both the TTA model and the reward model through the feedback mechanism,
something that KD or pseudo-label cannot achieve.
RLCF significantly outperforms the entropy minimization method --- TPT.
TPT can only learn from the TTA model itself and lacks awareness of the correctness of its predictions.
Figure~\ref{fig-ece} presents the expected calibration error~(ECE)~\citep{DBLP:conf/icml/GuoPSW17} of TPT and RLCF.
The ECE of the two both increases along with the TTA steps, but the ECE of RLCF is clearly lower.
This means the output of RLCF better reflects its uncertainty about the input and is more reliable.
In Figure~\ref{fig1:top-5-tpt-rlcf}\&\ref{fig:vis-cls-rlcf}, RLCF provides multiple positive scores for various objects, preventing the model from becoming blindly confident.
In Figure~\ref{fig-tpt-acc-steps}, the top-5 accuracy of TPT drops with more steps. The model is stuck in its incorrect predictions and pushes away the ground truth as shown in Figure~\ref{fig1:top-5-tpt-rlcf}.
By contrast, there is no such issue for RLCF in Figure~\ref{fig-rlcf-acc-steps}.

Ablation study about sampling factors and reward model choices can be found in Appendix~\ref{sec:app-abla}.



\subsection{Zero-shot text-image retrieval}
\textbf{Implementation details~~}
For text-image retrieval, we use the
test set of Flickr30K~\citep{plummer2015flickr30k}
and test split of MS-COCO~\citep{lin2014microsoft} divided by Karpathy~\textit{et al.}~\citep{karpathy2015deep}. 
Each image in the two test sets corresponds to 5 sentences.
CLIP-ViT-B/16 is adopted as the retrieval model.
The learning rate is 1e-6, the weight decay value is 5e-4, and AdamW optimizer is used.
For MS-COCO, the sampling factor $K=12$ for text-to-image retrieval, and $K=20$ for the other case.
For Flickr30K, $K=12$ and $K=16$ for text-to-image and image-to-text retrieval, respectively.
The adaptation steps are 8.
For the momentum buffer, $m=0.9998$ and $B_s=64$.
We also compare RLCF with knowledge distillation~(KD) with CLIP-ViT-L/14 as the teacher.

\textbf{Results~~}
Table~\ref{tab:retrieval} presents the retrieval results on MS-COCO
and Flickr30K. 
RLCF demonstrates significant improvement compared to the zero-shot baseline and even outperforms the most powerful CLIP-ViT-L/14-336.
Similar phenomena are also observed in zero-shot classification.
The feedback mechanism reserves the merits of the TTA model and makes the TTA model improve with the reward model.
In contrast, KD or pseudo-label forces the student to mimic the teacher regardless of the correctness of the teacher as discussed in Sec.~\ref{sec:task-tta}.
In KD for supervised classification~\citep{hinton2015distilling,zhao2022decoupled,wang2021accelerate}, the student is generally worse than the teacher due to their capacity gap and incomplete learning.
Nevertheless, RLCF can surpass the powerful reward model with the feedback mechanism during test time in a zero-shot circumstance.

\begin{table}[!t]
\vspace{-0.5cm}
	\caption
	{
\textbf{TTA for zero-shot text-image
    retrieval}. KD uses CLIP-ViT-L/14 as the teacher model.
      Improvement in Recall@1 with RLCF compared to the CLIP-ViT-B/16 baseline is in \textcolor{blue}{${(\uparrow}$blue)}.
	}
	\centering	
    \setlength\tabcolsep{4pt}
	\resizebox{0.96\textwidth}{!}{%
	\begin{tabular}	{l|cccccc|cccccc}
		\toprule	 	
	\multirow{3}{*}{Method} &
    \multicolumn{6}{c|}{MS-COCO (5K test images)} & \multicolumn{6}{c}{Flickr30K (1K test images)} \\
	
    & \multicolumn{3}{c}{text-to-image}
    & \multicolumn{3}{c|}{image-to-text}
    & \multicolumn{3}{c}{text-to-image}
    & \multicolumn{3}{c}{image-to-text} \\
	
    & R@1 &R@5 &R@10
    & R@1 &R@5 &R@10
    & R@1 &R@5 &R@10
    & R@1 &R@5 &R@10\\
    \midrule
    & \multicolumn{12}{c}{\textit{\underline{Zero-shot baseline}}}  \\

    CLIP-ViT-B/16 
    & 33.0	& 58.2	& 68.9
    & 52.5	& 76.8	& 84.6
    & 62.2	& 85.7	& 91.8
    & 81.2	& 96.4	& 98.5	\\

    CLIP-ViT-L/14 
    & 36.1	&60.9	& 71.1
    & 56.2	&78.9	& 86.9
    & 64.6	&87.1	& 92.1
    & 85.3	&97.2	& 99.1	\\

    CLIP-ViT-L/14-336
    & 36.6	& 60.9	& 71.0
    & 57.3	& 80.6	& 87.8
    & 67.1	& 88.9	& 93.2
    & 86.6	& 98.0	& 99.1	\\

    
    \midrule
    & \multicolumn{12}{c}{\underline{\textit{TTA for} CLIP-ViT-B/16}}  \\

    Pseudo-label~(\citeauthor{lee2013pseudo})
    & 33.0	& 57.8	& 68.2
    & 52.4	& 72.4	& 81.8
    & 62.2	& 85.3	& 91.7
    & 81.1	& 93.2	& 97.8	\\

    KD~(\citeauthor{hinton2015distilling})~(steps: 3) 
    & 37.6	& 61.0	& 70.7
    & 57.0	& 79.0	& 86.3
    & 66.9	& 87.9	& 92.9
    & 85.3	& \cellcolor{Gray2}97.5	& 98.5	\\

    KD~(\citeauthor{hinton2015distilling})~(steps: 5) 
    & 34.6	& 59.7	& 69.8
    & 53.7	& 76.6	& 84.5
    & 61.4	& 86.0	& 91.7
    & 83.1	& 95.9	& 97.9	\\

    \textbf{RLCF}
    & 37.3\textcolor{blue}{$_{(\uparrow4.3)}$}
    & 62.7	
    & 71.5
    & \cellcolor{Gray2}59.1\textcolor{blue}{$_{(\uparrow6.6)}$}
    & 80.1
    & 86.9
    & \cellcolor{Gray2}67.1\textcolor{blue}{$_{(\uparrow4.9)}$}
    & 89.1	
    & \cellcolor{Gray2}93.2
    & 87.3\textcolor{blue}{$_{(\uparrow6.1)}$}
    & 97.2	
    & \cellcolor{Gray2}98.8	\\

    \textbf{RLCF-S}
	& \cellcolor{Gray2}38.3\textcolor{blue}{$_{(\uparrow5.3)}$}
    & \cellcolor{Gray2}63.4	
    & \cellcolor{Gray2}72.5
    & \cellcolor{Gray1}60.8\textcolor{blue}{$_{(\uparrow8.3)}$}
    & \cellcolor{Gray1}80.8
    & \cellcolor{Gray2}87.5
    & \cellcolor{Gray1}68.5\textcolor{blue}{$_{(\uparrow6.3)}$}	
    & \cellcolor{Gray2}90.0	
    & \cellcolor{Gray1}93.7
    & \cellcolor{Gray1}88.3\textcolor{blue}{$_{(\uparrow7.1)}$}	
    & \cellcolor{Gray1}97.7	
    & \cellcolor{Gray1}98.9 \\

    \textbf{RLCF-S-M}
    &\cellcolor{Gray1}38.4\textcolor{blue}{$_{(\uparrow5.4)}$}
    & \cellcolor{Gray1}63.5
    & \cellcolor{Gray1}72.6
    & \cellcolor{Gray1}60.8\textcolor{blue}{$_{(\uparrow8.3)}$}	
    & \cellcolor{Gray2}80.5
    & \cellcolor{Gray1}87.6
    & \cellcolor{Gray1}68.5\textcolor{blue}{$_{(\uparrow6.3)}$}	
    & \cellcolor{Gray1}90.2	
    & \cellcolor{Gray1}93.7
    & \cellcolor{Gray2}88.1\textcolor{blue}{$_{(\uparrow6.9)}$}	
    & \cellcolor{Gray1}97.7	
    &\cellcolor{Gray1}98.9
	\\

	\bottomrule
	\end{tabular}}
    \label{tab:retrieval}
    \vspace{-0.3cm}
\end{table}		

\subsection{Image captioning}
\textbf{Datasets~~}
To test the adaptation ability of RLCF for captioning
models in a zero-shot or cross-domain condition, we train the
captioning model on MS-COCO train set~\citep{lin2014microsoft} and test it on the test set of Flickr30K~\citep{plummer2015flickr30k} and validation set of NoCaps~\citep{agrawal2019nocaps}.
NoCaps validation set contains three splits according to
whether contains MS-COCO objects: in domain contains only MS-COCO objects, near domain contains both MS-COCO and novel objects, and out domain contains only novel objects.

\textbf{Implementation details~~}
CLIPCap~\citep{mokady2021clipcap} and CapDec~\citep{nukrai2022text}, two LLM-based methods, are chosen as the captioning models.
The two have the same architecture, while CLIPCap is trained with CLIP-ViT-B/16 image embedding and CapDec is trained with CLIP-ViT-B/16 text embedding.
The projector in Figure~\ref{fig:tta-cap} is an 8-layer transformer encoder that contains about 43M parameters.
The LLM is an OPT-125M~\citep{zhang2022opt}.
During TTA, we only tune the parameters of the projector.
For CLIPCap, the learning rate is 2e-6, and sampling factor $K=10$; for CapDec, the learning rate is 5e-6 on Flickr30K, 3e-6 on NoCaps, and $K=6$. No weight decay is applied. The optimizer is AdamW. The TTA step is 4.
After TTA, captions are generated with a beam search with a
width of 5 and the final caption is the one with the highest score.

\textbf{Results~~} 
Table~\ref{tab:caption} presents results for image captioning. A weakly supervised method --- MAGIC~\citep{su2022language} and a zero-shot method ---DeCap~\citep{li2023decap}, are included for reference.
The reported metrics include BLEU@4, CIDEr, SPICE, and RefCLIPScore~\citep{hessel2021clipscore}. RefCLIPScore reflects the similarity between generated text and reference captions.
The improvements in CIDEr metric~\citep{vedantam2015cider} are highlighted.
For all metrics, both CapDec and CLIPCap with RLCF significantly improve upon the baselines.
This demonstrates the strong generalization ability of RLCF in image captioning, even with a single test sample.
It is noteworthy that CLIPCap achieves greater improvements in CIDEr~(up to 9.2) compared to CapDec.
CLIPCap can also use a large sampling factor $K$.
This is possible because CLIPCap can generate higher-quality candidate captions.
The results of RLCF-S-M are not shown as it is no
better than RLCF-S.

\begin{table}[!t]
\vspace{-0.5cm}
    \centering
    \caption{\textbf{TTA for image captioning}.
    B@4 for BLEU@4, C for CIDEr, S for SPICE,
    and Ref-C for {RefCLIPScore}.
    The gain of well-recognized CIDEr metric is in \textcolor{blue}{($\uparrow$blue)}.
    }
    \setlength\tabcolsep{4pt}
    \resizebox{0.96\textwidth}{!}{%
    \begin{tabular}{l|ccccccccc|ccccc}
    \toprule
    \multirow{3}{*}{Method}
    & \multicolumn{9}{c|}{MS-COCO $\Longrightarrow$ NoCaps }
    & \multicolumn{4}{c}{MS-COCO $\Longrightarrow$ Flickr30K } \\

    & \multicolumn{3}{c}{in domain}
    & \multicolumn{3}{c}{near domain}
    & \multicolumn{3}{c|}{out domain}
    & \multicolumn{4}{c}{Karpathy's test split} \\

    & B@4 & C & S
    & B@4 & C & S
    & B@4 & C & S
    & B@4 & C & S & Ref-C \\
    \midrule

    MAGIC~\citep{su2022language}
    & - & - & -
    & - & - & -
    & - & - & -
    & 5.2 & 18.3 & 5.7 & - \\

    DeCap~\citep{li2023decap}
    & - & 72.7 & -
    & - & 61.9 & -
    & - & 43.9 & -
    & 17.7 & 42.0 & 13.8 & - \\
    \midrule

    & \multicolumn{13}{c}{\underline{\textit{TTA for }CapDec~(\textbf{zero-shot})}} \\

    CapDec~\citep{nukrai2022text}
    & 32.4 & 62.6 & 10.3
    & 29.2 & 54.0 & 9.6
    & 17.2 & 31.7 & 6.4
    & 19.3 & 37.0 & 11.7 & 74.1 \\

    + \textbf{RLCF}
    & 33.3 & 68.0\textcolor{blue}{$_{(\uparrow5.3)}$} & 10.7
    & 30.3 & 57.9\textcolor{blue}{$_{(\uparrow3.9)}$}
    & 10.3
    & 17.6 
    & \cellcolor{Gray1}35.5\textcolor{blue}{$_{(\uparrow3.8)}$}
    & 6.9
    & \cellcolor{Gray1}20.3
    & \cellcolor{Gray1}41.9\textcolor{blue}{$_{(\uparrow4.9)}$}
    & \cellcolor{Gray1}12.7
    & \cellcolor{Gray1}75.7 \\

    + \textbf{RLCF-S}
    & \cellcolor{Gray1}34.0
    & \cellcolor{Gray1}68.3\textcolor{blue}{$_{(\uparrow5.7)}$}
    & \cellcolor{Gray1}10.8
    & \cellcolor{Gray1}30.3 
    & \cellcolor{Gray1}58.5\textcolor{blue}{$_{(\uparrow4.5)}$}
    & \cellcolor{Gray1}10.3
    & \cellcolor{Gray1}17.7
    & 35.3\textcolor{blue}{$_{(\uparrow3.6)}$}
    & \cellcolor{Gray1}6.9
    & 20.1
    & 41.6\textcolor{blue}{$_{(\uparrow4.6)}$}
    & 12.7 & 75.7 \\


    \midrule
    & \multicolumn{13}{c}{\underline{\textit{TTA for }CLIPCap~(\textbf{cross-domain})}} \\

    CLIPCap~\citep{mokady2021clipcap}
    & 36.3 & 76.9 & 11.9
    & 34.8 & 73.5 & 11.0
    & 22.5 & 54.6 & 8.6
    & 21.8 & 49.3 & 13.1 & 76.7 \\

    + \textbf{RLCF}
    & 38.6 
    & 84.0\textcolor{blue}{$_{(\uparrow7.1)}$}
    & 12.5
    & \cellcolor{Gray1}36.1 
    & 79.6\textcolor{blue}{$_{(\uparrow6.1)}$}
    & 11.8
    & \cellcolor{Gray1}24.7
    & \cellcolor{Gray1}63.8\textcolor{blue}{$_{(\uparrow9.2)}$}
    & \cellcolor{Gray1}9.6
    & 23.3 & 56.6\textcolor{blue}{$_{(\uparrow7.3)}$}
    & 14.5 & 79.4 \\

    + \textbf{RLCF-S} 
    & \cellcolor{Gray1}38.7 
    & \cellcolor{Gray1}84.7\textcolor{blue}{$_{(\uparrow7.8)}$}
    & \cellcolor{Gray1}12.6
    & 35.8 
    & \cellcolor{Gray1}79.7\textcolor{blue}{$_{(\uparrow6.2)}$}
    & \cellcolor{Gray1}11.8 
    & 24.2
    & 63.1\textcolor{blue}{$_{(\uparrow8.5)}$} 
    & 9.5
    & \cellcolor{Gray1}23.6 
    & \cellcolor{Gray1}57.8\textcolor{blue}{$_{(\uparrow8.5)}$}
    & \cellcolor{Gray1}14.6 
    & \cellcolor{Gray1}79.4 \\
    
    \bottomrule
    \end{tabular}}
\vspace{-0.1cm}
\label{tab:caption} 
\end{table}

\begin{figure}[!t]
\centering
\includegraphics[width=0.96\linewidth]{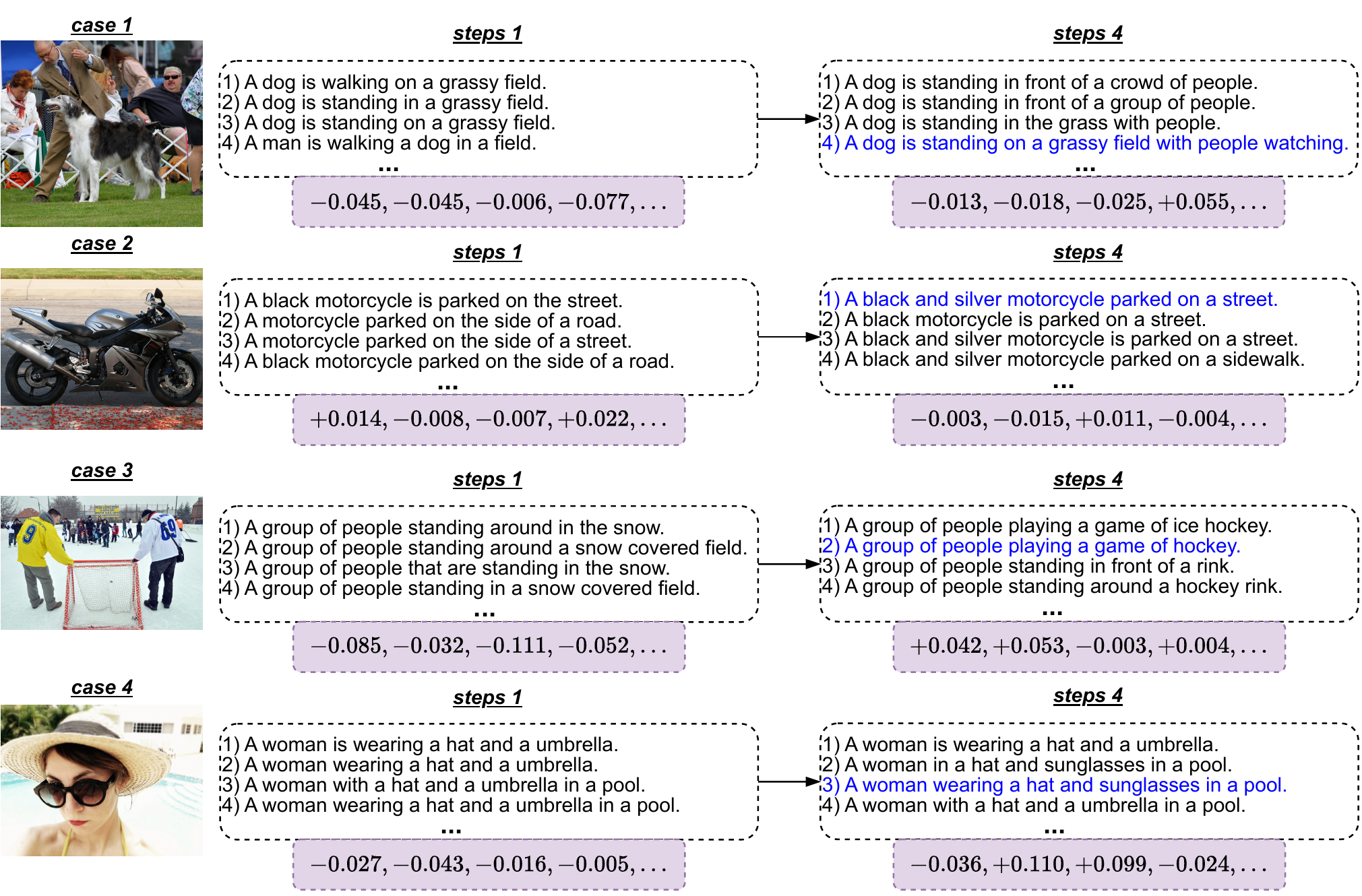}
\caption{
\textbf{Intermediate generated captions of CLIPCap and the CLIP reward}.
The sampling factor $K=10$, only 4 candidates are shown here.
The final generated caption is in \textcolor{blue}{blue}.
}
\label{fig:tta-cap-vis}
\vspace{-0.3cm}
\end{figure}

\textbf{Qualitative results~~}
Figure~\ref{fig:tta-cap-vis} displays the intermediate-generated captions and their corresponding rewards.
The visualization reveals that the CLIP reward model favors captions that provide a holistic description of the image.
Through feedback, the generation of such captions is encouraged.
During TTA, captions aligned with the preferences of CLIP are given higher priority.
Please refer to Figure~\ref{fig:vis-cap-rlcf} in Appendix~\ref{sec:app-vis} for more visualization cases.

\section{Conclusion}
\label{sec:conclusion}
In this work, we introduce reinforcement learning with CLIP feedback~(RLCF) to improve the zero-shot generalization ability of VLMs on the fly.
A novel reward function with CLIP is developed.
We instantiate three TTA pipelines for image classification, text-image retrieval, and image captioning with task-specific sampling strategies and parameter tuning manners.
With RLCF, the zero-shot generalization capacity of various VLMs is boosted significantly.
We hope RLCF can provide heuristic information for future research that employs TTA with feedback from large foundation models.

\pagebreak

\section*{Acknowledgments}
This work was supported in part by the Australian Research Council (ARC) under Grant DP200100938.
Thanks Chao Liang for his helpful discussions.

\bibliography{iclr2024_conference}

\begin{thebibliography}{62}
\providecommand{\natexlab}[1]{#1}
\providecommand{\url}[1]{\texttt{#1}}
\expandafter\ifx\csname urlstyle\endcsname\relax
  \providecommand{\doi}[1]{doi: #1}\else
  \providecommand{\doi}{doi: \begingroup \urlstyle{rm}\Url}\fi

\bibitem[Agrawal et~al.(2019)Agrawal, Desai, Wang, Chen, Jain, Johnson, Batra, Parikh, Lee, and Anderson]{agrawal2019nocaps}
Harsh Agrawal, Karan Desai, Yufei Wang, Xinlei Chen, Rishabh Jain, Mark Johnson, Dhruv Batra, Devi Parikh, Stefan Lee, and Peter Anderson.
\newblock Nocaps: Novel object captioning at scale.
\newblock In \emph{ICCV}, 2019.

\bibitem[Bai et~al.(2022)Bai, Jones, Ndousse, Askell, Chen, DasSarma, Drain, Fort, Ganguli, Henighan, et~al.]{bai2022training}
Yuntao Bai, Andy Jones, Kamal Ndousse, Amanda Askell, Anna Chen, Nova DasSarma, Dawn Drain, Stanislav Fort, Deep Ganguli, Tom Henighan, et~al.
\newblock Training a helpful and harmless assistant with reinforcement learning from human feedback.
\newblock \emph{arXiv preprint arXiv:2204.05862}, 2022.

\bibitem[Chen et~al.(2020)Chen, Kornblith, Norouzi, and Hinton]{chen2020simple}
Ting Chen, Simon Kornblith, Mohammad Norouzi, and Geoffrey Hinton.
\newblock A simple framework for contrastive learning of visual representations.
\newblock In \emph{ICML}, 2020.

\bibitem[Cho et~al.(2022)Cho, Yoon, Kale, Dernoncourt, Bui, and Bansal]{Cho2022CLIPReward}
Jaemin Cho, Seunghyun Yoon, Ajinkya Kale, Franck Dernoncourt, Trung Bui, and Mohit Bansal.
\newblock Fine-grained image captioning with clip reward.
\newblock In \emph{Findings of NAACL}, 2022.

\bibitem[Deng et~al.(2009)Deng, Dong, Socher, Li, Li, and Fei-Fei]{imagenet_cvpr09}
J.~Deng, W.~Dong, R.~Socher, L.-J. Li, K.~Li, and L.~Fei-Fei.
\newblock {ImageNet: A Large-Scale Hierarchical Image Database}.
\newblock In \emph{CVPR}, 2009.

\bibitem[Deng et~al.(2022)Deng, Wang, Hsieh, Wang, Guo, Shu, Song, Xing, and Hu]{deng-etal-2022-rlprompt}
Mingkai Deng, Jianyu Wang, Cheng-Ping Hsieh, Yihan Wang, Han Guo, Tianmin Shu, Meng Song, Eric Xing, and Zhiting Hu.
\newblock {RLP}rompt: Optimizing discrete text prompts with reinforcement learning.
\newblock In \emph{EMNLP}, 2022.
\newblock URL \url{https://aclanthology.org/2022.emnlp-main.222}.

\bibitem[Dosovitskiy et~al.(2021)Dosovitskiy, Beyer, Kolesnikov, Weissenborn, Zhai, Unterthiner, Dehghani, Minderer, Heigold, Gelly, Uszkoreit, and Houlsby]{dosovitskiy2021an}
Alexey Dosovitskiy, Lucas Beyer, Alexander Kolesnikov, Dirk Weissenborn, Xiaohua Zhai, Thomas Unterthiner, Mostafa Dehghani, Matthias Minderer, Georg Heigold, Sylvain Gelly, Jakob Uszkoreit, and Neil Houlsby.
\newblock An image is worth 16x16 words: Transformers for image recognition at scale.
\newblock In \emph{ICLR}, 2021.
\newblock URL \url{https://openreview.net/forum?id=YicbFdNTTy}.

\bibitem[Glaese et~al.(2022)Glaese, McAleese, Tr{\k{e}}bacz, Aslanides, Firoiu, Ewalds, Rauh, Weidinger, Chadwick, Thacker, et~al.]{glaese2022improving}
Amelia Glaese, Nat McAleese, Maja Tr{\k{e}}bacz, John Aslanides, Vlad Firoiu, Timo Ewalds, Maribeth Rauh, Laura Weidinger, Martin Chadwick, Phoebe Thacker, et~al.
\newblock Improving alignment of dialogue agents via targeted human judgements.
\newblock \emph{arXiv preprint arXiv:2209.14375}, 2022.

\bibitem[Guo et~al.(2017)Guo, Pleiss, Sun, and Weinberger]{DBLP:conf/icml/GuoPSW17}
Chuan Guo, Geoff Pleiss, Yu~Sun, and Kilian~Q. Weinberger.
\newblock On calibration of modern neural networks.
\newblock In Doina Precup and Yee~Whye Teh (eds.), \emph{ICML}, 2017.

\bibitem[Guo et~al.(2020)Guo, Chen, Hu, Zhu, He, and Cai]{guo2020reducing}
Jia Guo, Minghao Chen, Yao Hu, Chen Zhu, Xiaofei He, and Deng Cai.
\newblock Reducing the teacher-student gap via spherical knowledge disitllation.
\newblock \emph{arXiv preprint arXiv:2010.07485}, 2020.

\bibitem[He et~al.(2016)He, Zhang, Ren, and Sun]{resnet}
Kaiming He, Xiangyu Zhang, Shaoqing Ren, and Jian Sun.
\newblock Deep residual learning for image recognition.
\newblock In \emph{CVPR}, 2016.

\bibitem[Hendrycks et~al.(2021{\natexlab{a}})Hendrycks, Basart, Mu, Kadavath, Wang, Dorundo, Desai, Zhu, Parajuli, Guo, Song, Steinhardt, and Gilmer]{Hendrycks2021imr}
Dan Hendrycks, Steven Basart, Norman Mu, Saurav Kadavath, Frank Wang, Evan Dorundo, Rahul Desai, Tyler Zhu, Samyak Parajuli, Mike Guo, Dawn Song, Jacob Steinhardt, and Justin Gilmer.
\newblock The many faces of robustness: {A} critical analysis of out-of-distribution generalization.
\newblock In \emph{ICCV}, 2021{\natexlab{a}}.

\bibitem[Hendrycks et~al.(2021{\natexlab{b}})Hendrycks, Zhao, Basart, Steinhardt, and Song]{Hendrycks2021ima}
Dan Hendrycks, Kevin Zhao, Steven Basart, Jacob Steinhardt, and Dawn Song.
\newblock Natural adversarial examples.
\newblock In \emph{CVPR}, pp.\  15262--15271, 2021{\natexlab{b}}.

\bibitem[Hessel et~al.(2021)Hessel, Holtzman, Forbes, Bras, and Choi]{hessel2021clipscore}
Jack Hessel, Ari Holtzman, Maxwell Forbes, Ronan~Le Bras, and Yejin Choi.
\newblock {CLIPScore:} a reference-free evaluation metric for image captioning.
\newblock In \emph{EMNLP}, 2021.

\bibitem[Hinton et~al.(2015)Hinton, Vinyals, and Dean]{hinton2015distilling}
Geoffrey Hinton, Oriol Vinyals, and Jeff Dean.
\newblock Distilling the knowledge in a neural network.
\newblock \emph{arXiv preprint arXiv:1503.02531}, 2015.

\bibitem[Hong et~al.(2022)Hong, Zhang, Pan, Cai, Yang, and Liu]{HongZPCYL22}
Fangzhou Hong, Mingyuan Zhang, Liang Pan, Zhongang Cai, Lei Yang, and Ziwei Liu.
\newblock Avatarclip: zero-shot text-driven generation and animation of 3d avatars.
\newblock \emph{{ACM} Trans. Graph.}, 2022.

\bibitem[Jia et~al.(2021)Jia, Yang, Xia, Chen, Parekh, Pham, Le, Sung, Li, and Duerig]{jia2021align}
Chao Jia, Yinfei Yang, Ye~Xia, Yi{-}Ting Chen, Zarana Parekh, Hieu Pham, Quoc~V. Le, Yun{-}Hsuan Sung, Zhen Li, and Tom Duerig.
\newblock Scaling up visual and vision-language representation learning with noisy text supervision.
\newblock In \emph{ICML}, 2021.

\bibitem[Karpathy \& Fei-Fei(2015)Karpathy and Fei-Fei]{karpathy2015deep}
Andrej Karpathy and Li~Fei-Fei.
\newblock Deep visual-semantic alignments for generating image descriptions.
\newblock In \emph{CVPR}, 2015.

\bibitem[Le et~al.(2022)Le, Rathour, Yamazaki, Luu, and Savvides]{le2022deep}
Ngan Le, Vidhiwar~Singh Rathour, Kashu Yamazaki, Khoa Luu, and Marios Savvides.
\newblock Deep reinforcement learning in computer vision: a comprehensive survey.
\newblock \emph{Artificial Intelligence Review}, 2022.

\bibitem[Lee et~al.(2013)]{lee2013pseudo}
Dong-Hyun Lee et~al.
\newblock Pseudo-label: The simple and efficient semi-supervised learning method for deep neural networks.
\newblock In \emph{Workshop on challenges in representation learning, ICML}, 2013.

\bibitem[Lee et~al.(2023)Lee, Phatale, Mansoor, Lu, Mesnard, Bishop, Carbune, and Rastogi]{lee2023rlaif}
Harrison Lee, Samrat Phatale, Hassan Mansoor, Kellie Lu, Thomas Mesnard, Colton Bishop, Victor Carbune, and Abhinav Rastogi.
\newblock Rlaif: Scaling reinforcement learning from human feedback with ai feedback, 2023.

\bibitem[Li et~al.(2023)Li, Zhu, Wen, and Yang]{li2023decap}
Wei Li, Linchao Zhu, Longyin Wen, and Yi~Yang.
\newblock Decap: Decoding {CLIP} latents for zero-shot captioning via text-only training.
\newblock In \emph{ICLR}, 2023.
\newblock URL \url{https://openreview.net/forum?id=Lt8bMlhiwx2}.

\bibitem[Lin et~al.(2014)Lin, Maire, Belongie, Hays, Perona, Ramanan, Doll{\'a}r, and Zitnick]{lin2014microsoft}
Tsung-Yi Lin, Michael Maire, Serge Belongie, James Hays, Pietro Perona, Deva Ramanan, Piotr Doll{\'a}r, and C~Lawrence Zitnick.
\newblock Microsoft coco: Common objects in context.
\newblock In \emph{ECCV}, 2014.

\bibitem[Lin et~al.(2023)Lin, Mirza, Kozinski, Possegger, Kuehne, and Bischof]{lin2022video}
Wei Lin, Muhammad~Jehanzeb Mirza, Mateusz Kozinski, Horst Possegger, Hilde Kuehne, and Horst Bischof.
\newblock Video test-time adaptation for action recognition.
\newblock \emph{CVPR}, 2023.

\bibitem[Liu et~al.(2021)Liu, Kothari, Van~Delft, Bellot-Gurlet, Mordan, and Alahi]{liu2021ttt++}
Yuejiang Liu, Parth Kothari, Bastien Van~Delft, Baptiste Bellot-Gurlet, Taylor Mordan, and Alexandre Alahi.
\newblock Ttt++: When does self-supervised test-time training fail or thrive?
\newblock \emph{NeurIPS}, 2021.

\bibitem[Loshchilov \& Hutter(2019)Loshchilov and Hutter]{loshchilov2018decoupled}
Ilya Loshchilov and Frank Hutter.
\newblock Decoupled weight decay regularization.
\newblock In \emph{ICLR}, 2019.
\newblock URL \url{https://openreview.net/forum?id=Bkg6RiCqY7}.

\bibitem[Manli et~al.(2022)Manli, Weili, De-An, Zhiding, Tom, Anima, and Chaowei]{shu2022tpt}
Shu Manli, Nie Weili, Huang De-An, Yu~Zhiding, Goldstein Tom, Anandkumar Anima, and Xiao Chaowei.
\newblock Test-time prompt tuning for zero-shot generalization in vision-language models.
\newblock In \emph{NeurIPS}, 2022.

\bibitem[Minderer et~al.(2021)Minderer, Djolonga, Romijnders, Hubis, Zhai, Houlsby, Tran, and Lucic]{DBLP:conf/nips/MindererDRHZHTL21}
Matthias Minderer, Josip Djolonga, Rob Romijnders, Frances Hubis, Xiaohua Zhai, Neil Houlsby, Dustin Tran, and Mario Lucic.
\newblock Revisiting the calibration of modern neural networks.
\newblock In \emph{NeurIPS}, 2021.

\bibitem[Mokady et~al.(2021)Mokady, Hertz, and Bermano]{mokady2021clipcap}
Ron Mokady, Amir Hertz, and Amit~H Bermano.
\newblock Clipcap: Clip prefix for image captioning.
\newblock \emph{arXiv preprint arXiv:2111.09734}, 2021.

\bibitem[Niu et~al.(2022)Niu, Wu, Zhang, Chen, Zheng, Zhao, and Tan]{niu2022efficient}
Shuaicheng Niu, Jiaxiang Wu, Yifan Zhang, Yaofo Chen, Shijian Zheng, Peilin Zhao, and Mingkui Tan.
\newblock Efficient test-time model adaptation without forgetting.
\newblock In \emph{ICML}, 2022.

\bibitem[Niu et~al.(2023)Niu, Wu, Zhang, Wen, Chen, Zhao, and Tan]{niu2023towards}
Shuaicheng Niu, Jiaxiang Wu, Yifan Zhang, Zhiquan Wen, Yaofo Chen, Peilin Zhao, and Mingkui Tan.
\newblock Towards stable test-time adaptation in dynamic wild world.
\newblock \emph{ICLR}, 2023.

\bibitem[Nukrai et~al.(2022)Nukrai, Mokady, and Globerson]{nukrai2022text}
David Nukrai, Ron Mokady, and Amir Globerson.
\newblock Text-only training for image captioning using noise-injected clip.
\newblock In \emph{Findings of EMNLP}, 2022.

\bibitem[OpenAI(2023)]{openai2023gpt}
OpenAI.
\newblock Gpt-4 technical report.
\newblock \emph{arXiv}, 2023.

\bibitem[Ouyang et~al.(2022)Ouyang, Wu, Jiang, Almeida, Wainwright, Mishkin, Zhang, Agarwal, Slama, Ray, et~al.]{ouyang2022training}
Long Ouyang, Jeffrey Wu, Xu~Jiang, Diogo Almeida, Carroll Wainwright, Pamela Mishkin, Chong Zhang, Sandhini Agarwal, Katarina Slama, Alex Ray, et~al.
\newblock Training language models to follow instructions with human feedback.
\newblock \emph{NeurIPS}, 2022.

\bibitem[Pinto et~al.(2023)Pinto, Kolesnikov, Shi, Beyer, and Zhai]{pinto2023tuning}
Andr{\'e}~Susano Pinto, Alexander Kolesnikov, Yuge Shi, Lucas Beyer, and Xiaohua Zhai.
\newblock Tuning computer vision models with task rewards.
\newblock \emph{arXiv preprint arXiv:2302.08242}, 2023.

\bibitem[Plummer et~al.(2015)Plummer, Wang, Cervantes, Caicedo, Hockenmaier, and Lazebnik]{plummer2015flickr30k}
Bryan~A Plummer, Liwei Wang, Chris~M Cervantes, Juan~C Caicedo, Julia Hockenmaier, and Svetlana Lazebnik.
\newblock Flickr30k entities: Collecting region-to-phrase correspondences for richer image-to-sentence models.
\newblock In \emph{ICCV}, 2015.

\bibitem[Radford et~al.(2021)Radford, Kim, Hallacy, Ramesh, Goh, Agarwal, Sastry, Askell, Mishkin, Clark, Krueger, and Sutskever]{radford2021clip}
Alec Radford, Jong~Wook Kim, Chris Hallacy, Aditya Ramesh, Gabriel Goh, Sandhini Agarwal, Girish Sastry, Amanda Askell, Pamela Mishkin, Jack Clark, Gretchen Krueger, and Ilya Sutskever.
\newblock Learning transferable visual models from natural language supervision.
\newblock In \emph{ICML}, 2021.

\bibitem[Recht et~al.(2019)Recht, Roelofs, Schmidt, and Shankar]{Recht2019imv2}
Benjamin Recht, Rebecca Roelofs, Ludwig Schmidt, and Vaishaal Shankar.
\newblock Do imagenet classifiers generalize to imagenet?
\newblock In Kamalika Chaudhuri and Ruslan Salakhutdinov (eds.), \emph{ICML}, 2019.

\bibitem[Rennie et~al.(2017)Rennie, Marcheret, Mroueh, Ross, and Goel]{rennie2017self}
Steven~J Rennie, Etienne Marcheret, Youssef Mroueh, Jerret Ross, and Vaibhava Goel.
\newblock Self-critical sequence training for image captioning.
\newblock In \emph{CVPR}, 2017.

\bibitem[Sain et~al.(2023)Sain, Bhunia, Chowdhury, Koley, Xiang, and Song]{DBLP:conf/cvpr/SainBCKXS23}
Aneeshan Sain, Ayan~Kumar Bhunia, Pinaki~Nath Chowdhury, Subhadeep Koley, Tao Xiang, and Yi{-}Zhe Song.
\newblock {CLIP} for all things zero-shot sketch-based image retrieval, fine-grained or not.
\newblock In \emph{CVPR}, 2023.

\bibitem[Schneider et~al.(2020)Schneider, Rusak, Eck, Bringmann, Brendel, and Bethge]{schneider2020improving}
Steffen Schneider, Evgenia Rusak, Luisa Eck, Oliver Bringmann, Wieland Brendel, and Matthias Bethge.
\newblock Improving robustness against common corruptions by covariate shift adaptation.
\newblock \emph{NeurIPS}, 2020.

\bibitem[Schulman et~al.(2017)Schulman, Wolski, Dhariwal, Radford, and Klimov]{schulman2017proximal}
John Schulman, Filip Wolski, Prafulla Dhariwal, Alec Radford, and Oleg Klimov.
\newblock Proximal policy optimization algorithms.
\newblock \emph{arXiv preprint arXiv:1707.06347}, 2017.

\bibitem[Stiennon et~al.(2020)Stiennon, Ouyang, Wu, Ziegler, Lowe, Voss, Radford, Amodei, and Christiano]{stiennon2020learning}
Nisan Stiennon, Long Ouyang, Jeffrey Wu, Daniel Ziegler, Ryan Lowe, Chelsea Voss, Alec Radford, Dario Amodei, and Paul~F Christiano.
\newblock Learning to summarize with human feedback.
\newblock \emph{NeurIPS}, 2020.

\bibitem[Su et~al.(2022)Su, Lan, Liu, Liu, Yogatama, Wang, Kong, and Collier]{su2022language}
Yixuan Su, Tian Lan, Yahui Liu, Fangyu Liu, Dani Yogatama, Yan Wang, Lingpeng Kong, and Nigel Collier.
\newblock Language models can see: plugging visual controls in text generation.
\newblock \emph{arXiv preprint arXiv:2205.02655}, 2022.

\bibitem[Sun et~al.(2020)Sun, Wang, Liu, Miller, Efros, and Hardt]{sun2020ttt}
Yu~Sun, Xiaolong Wang, Zhuang Liu, John Miller, Alexei~A. Efros, and Moritz Hardt.
\newblock Test-time training with self-supervision for generalization under distribution shifts.
\newblock In \emph{ICML}, 2020.

\bibitem[Vaswani et~al.(2017)Vaswani, Shazeer, Parmar, Uszkoreit, Jones, Gomez, Kaiser, and Polosukhin]{vaswani2017attention}
Ashish Vaswani, Noam Shazeer, Niki Parmar, Jakob Uszkoreit, Llion Jones, Aidan~N Gomez, {\L}ukasz Kaiser, and Illia Polosukhin.
\newblock Attention is all you need.
\newblock \emph{NeurIPS}, 2017.

\bibitem[Vedantam et~al.(2015)Vedantam, Lawrence~Zitnick, and Parikh]{vedantam2015cider}
Ramakrishna Vedantam, C~Lawrence~Zitnick, and Devi Parikh.
\newblock Cider: Consensus-based image description evaluation.
\newblock In \emph{CVPR}, 2015.

\bibitem[Wang et~al.(2021{\natexlab{a}})Wang, Shelhamer, Liu, Olshausen, and Darrell]{wang2021tent}
Dequan Wang, Evan Shelhamer, Shaoteng Liu, Bruno~A. Olshausen, and Trevor Darrell.
\newblock Tent: Fully test-time adaptation by entropy minimization.
\newblock In \emph{ICLR}, 2021{\natexlab{a}}.

\bibitem[Wang et~al.(2019)Wang, Ge, Lipton, and Xing]{wang2019learning}
Haohan Wang, Songwei Ge, Zachary Lipton, and Eric~P Xing.
\newblock Learning robust global representations by penalizing local predictive power.
\newblock In \emph{NeurIPS}, 2019.

\bibitem[Wang et~al.(2021{\natexlab{b}})Wang, Chen, Zhao, Chen, Hu, Liu, Cai, He, and Liu]{wang2021accelerate}
Wenxiao Wang, Minghao Chen, Shuai Zhao, Long Chen, Jinming Hu, Haifeng Liu, Deng Cai, Xiaofei He, and Wei Liu.
\newblock Accelerate cnns from three dimensions: A comprehensive pruning framework.
\newblock In \emph{ICML}, 2021{\natexlab{b}}.

\bibitem[Williams(1992)]{williams1992reinforce}
Ronald~J Williams.
\newblock Simple statistical gradient-following algorithms for connectionist reinforcement learning.
\newblock \emph{Machine learning}, 1992.

\bibitem[Wortsman et~al.(2022)Wortsman, Ilharco, Gadre, Roelofs, Gontijo-Lopes, Morcos, Namkoong, Farhadi, Carmon, Kornblith, et~al.]{wortsman2022model}
Mitchell Wortsman, Gabriel Ilharco, Samir~Ya Gadre, Rebecca Roelofs, Raphael Gontijo-Lopes, Ari~S Morcos, Hongseok Namkoong, Ali Farhadi, Yair Carmon, Simon Kornblith, et~al.
\newblock Model soups: averaging weights of multiple fine-tuned models improves accuracy without increasing inference time.
\newblock In \emph{ICML}, 2022.

\bibitem[Xu et~al.(2023)Xu, Liu, Wu, Tong, Li, Ding, Tang, and Dong]{xu2023imagereward}
Jiazheng Xu, Xiao Liu, Yuchen Wu, Yuxuan Tong, Qinkai Li, Ming Ding, Jie Tang, and Yuxiao Dong.
\newblock Imagereward: Learning and evaluating human preferences for text-to-image generation.
\newblock \emph{arXiv preprint arXiv:2304.05977}, 2023.

\bibitem[Yuan et~al.(2021)Yuan, Chen, Chen, Codella, Dai, Gao, Hu, Huang, Li, Li, Liu, Liu, Liu, Lu, Shi, Wang, Wang, Xiao, Xiao, Yang, Zeng, Zhou, and Zhang]{yuan2021florence}
Lu~Yuan, Dongdong Chen, Yi-Ling Chen, Noel Codella, Xiyang Dai, Jianfeng Gao, Houdong Hu, Xuedong Huang, Boxin Li, Chunyuan Li, Ce~Liu, Mengchen Liu, Zicheng Liu, Yumao Lu, Yu~Shi, Lijuan Wang, Jianfeng Wang, Bin Xiao, Zhen Xiao, Jianwei Yang, Michael Zeng, Luowei Zhou, and Pengchuan Zhang.
\newblock Florence: A new foundation model for computer vision.
\newblock \emph{arXiv preprint arXiv:2111.11432}, November 2021.

\bibitem[Zancato et~al.(2023)Zancato, Achille, Liu, Trager, Perera, and Soatto]{zancato2023train}
Luca Zancato, Alessandro Achille, Tian~Yu Liu, Matthew Trager, Pramuditha Perera, and Stefano Soatto.
\newblock Train/test-time adaptation with retrieval.
\newblock \emph{CVPR}, 2023.

\bibitem[Zhang et~al.(2022{\natexlab{a}})Zhang, Levine, and Finn]{zhang2022memo}
Marvin Zhang, Sergey Levine, and Chelsea Finn.
\newblock Memo: Test time robustness via adaptation and augmentation.
\newblock \emph{NeurIPS}, 2022{\natexlab{a}}.

\bibitem[Zhang et~al.(2022{\natexlab{b}})Zhang, Roller, Goyal, Artetxe, Chen, Chen, Dewan, Diab, Li, Lin, et~al.]{zhang2022opt}
Susan Zhang, Stephen Roller, Naman Goyal, Mikel Artetxe, Moya Chen, Shuohui Chen, Christopher Dewan, Mona Diab, Xian Li, Xi~Victoria Lin, et~al.
\newblock Opt: Open pre-trained transformer language models.
\newblock \emph{arXiv preprint arXiv:2205.01068}, 2022{\natexlab{b}}.

\bibitem[Zhang et~al.(2023)Zhang, Wang, Zhou, Schuurmans, and Gonzalez]{zhang2023tempera}
Tianjun Zhang, Xuezhi Wang, Denny Zhou, Dale Schuurmans, and Joseph~E Gonzalez.
\newblock Tempera: Test-time prompt editing via reinforcement learning.
\newblock In \emph{ICLR}, 2023.

\bibitem[Zhao et~al.(2022)Zhao, Cui, Song, Qiu, and Liang]{zhao2022decoupled}
Borui Zhao, Quan Cui, Renjie Song, Yiyu Qiu, and Jiajun Liang.
\newblock Decoupled knowledge distillation.
\newblock In \emph{CVPR}, 2022.

\bibitem[Zhou et~al.(2021)Zhou, Yang, Loy, and Liu]{zhou2021coop}
Kaiyang Zhou, Jingkang Yang, Chen~Change Loy, and Ziwei Liu.
\newblock Learning to prompt for vision-language models.
\newblock \emph{CoRR}, abs/2109.01134, 2021.

\bibitem[Zhou et~al.(2022)Zhou, Yang, Loy, and Liu]{zhou2022cocoop}
Kaiyang Zhou, Jingkang Yang, Chen~Change Loy, and Ziwei Liu.
\newblock Conditional prompt learning for vision-language models.
\newblock In \emph{CVPR}, 2022.

\bibitem[Ziegler et~al.(2019)Ziegler, Stiennon, Wu, Brown, Radford, Amodei, Christiano, and Irving]{ziegler2019fine}
Daniel~M Ziegler, Nisan Stiennon, Jeffrey Wu, Tom~B Brown, Alec Radford, Dario Amodei, Paul Christiano, and Geoffrey Irving.
\newblock Fine-tuning language models from human preferences.
\newblock \emph{arXiv preprint arXiv:1909.08593}, 2019.

\end{thebibliography}
\bibliographystyle{iclr2024_conference}

\newpage
\appendix
\section*{Appendix}
\section{Visualization results} \label{sec:app-vis}

\begin{figure*}[h]
\vspace{-0.25cm}
\centering
\includegraphics[width=\linewidth]{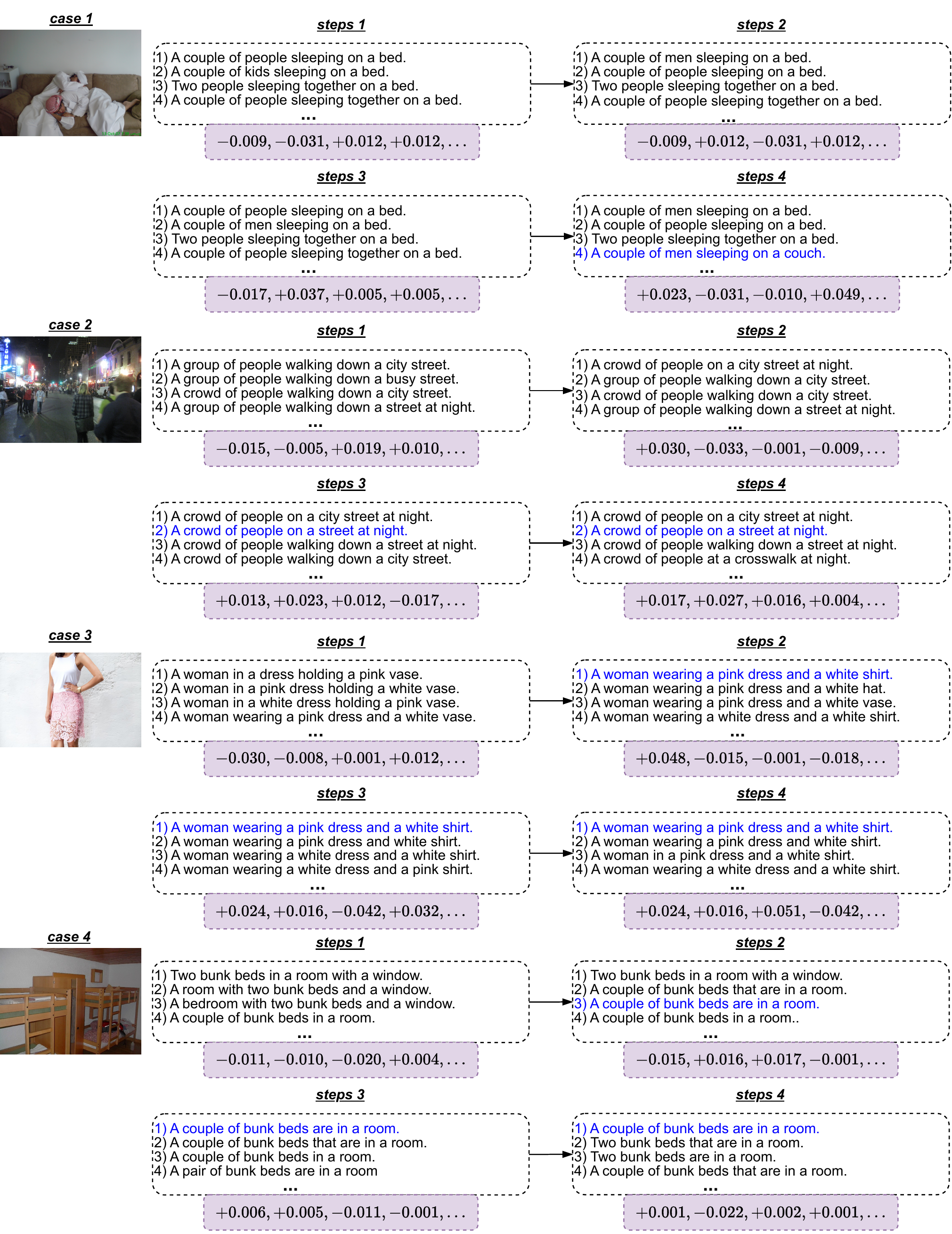}
\caption{
\textbf{Generated captions at each TTA step of CLIPCap and the CLIP reward}.
The sampling factor $K=10$, only 4 candidates are shown here.
The final generated caption is in \textcolor{blue}{blue}.
}
\label{fig:vis-cap-rlcf}
\end{figure*}

\begin{figure*}[t]
\vspace{-0.5cm}
\centering
\includegraphics[width=\linewidth]{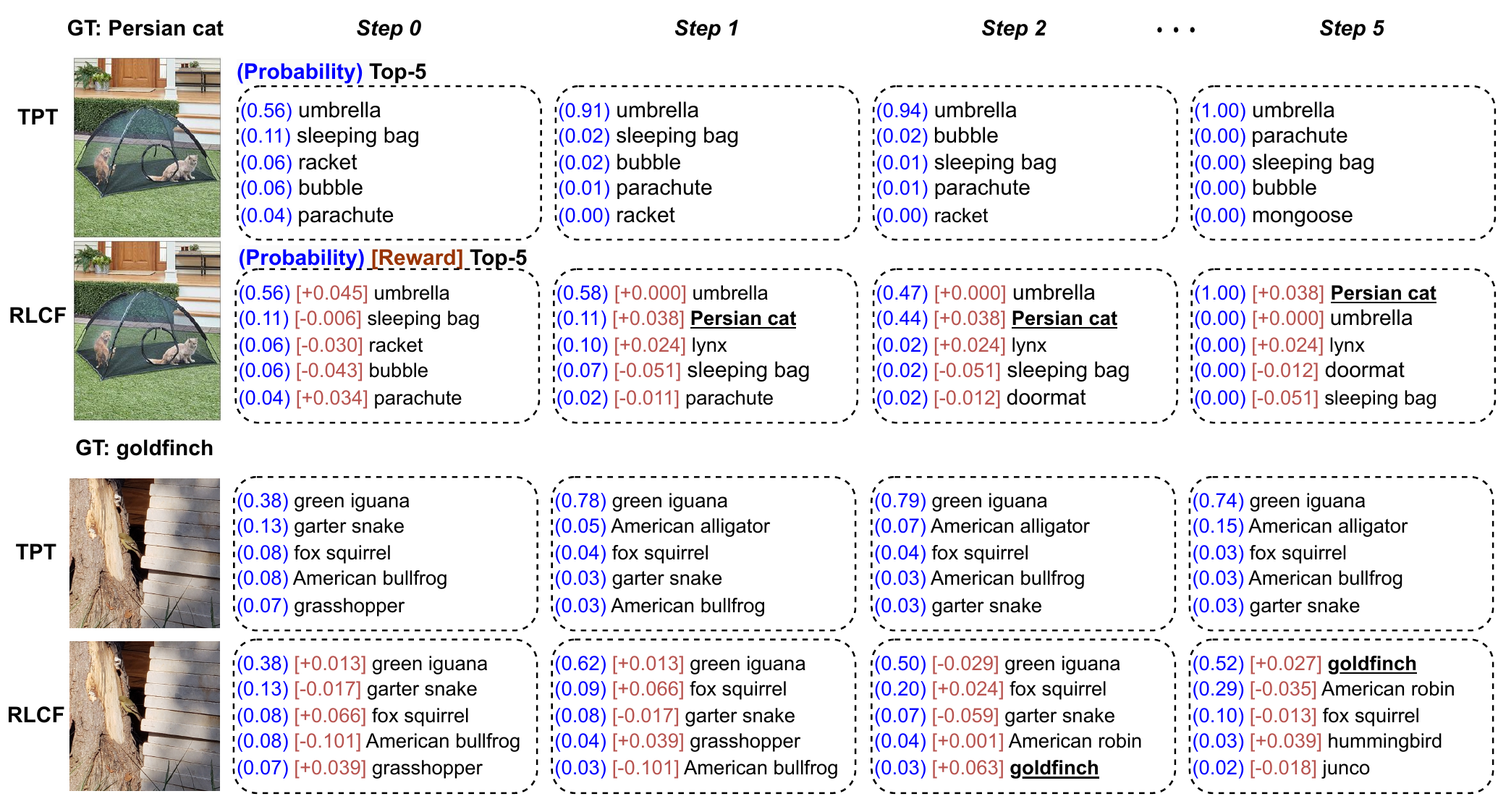}
	\caption{ \textbf{Top-5 predictions, probabilities, and rewards with different TTA steps}. Prompt tuning for CLIP-ViT-B/16. For RLCF, sampling factor $K=5$. We perform 5 TTA steps and step 0 means the original prediction. Images from ImageNet-A.}
\label{fig:vis-cls-rlcf}
\end{figure*}


In Figure~\ref{fig:vis-cap-rlcf}, we show more captioning examples like that in Figure~\ref{fig:tta-cap-vis}. These samples illustrate how CLIP reward helps the captioning model select a description that matches the picture more closely.

In Figure~\ref{fig:vis-cls-rlcf}, we provide visualization of top-5 predictions with different TTA steps like Figure~\ref{fig1:top-5-tpt-rlcf}.
In Figure~\ref{fig:vis-cls-rlcf}, the ground truth is not in the top-5 predictions. In such cases, TPT cannot find the object by minimizing the entropy of model outputs.
By contrast, RLCF can discover the ground truth by pushing away the objects with negative feedback.
\section{Ablation study}~\label{sec:app-abla}
\subsection{Sampling factor}


\begin{wrapfigure}{r}{6.6cm}
\centering
\includegraphics[width=\linewidth]{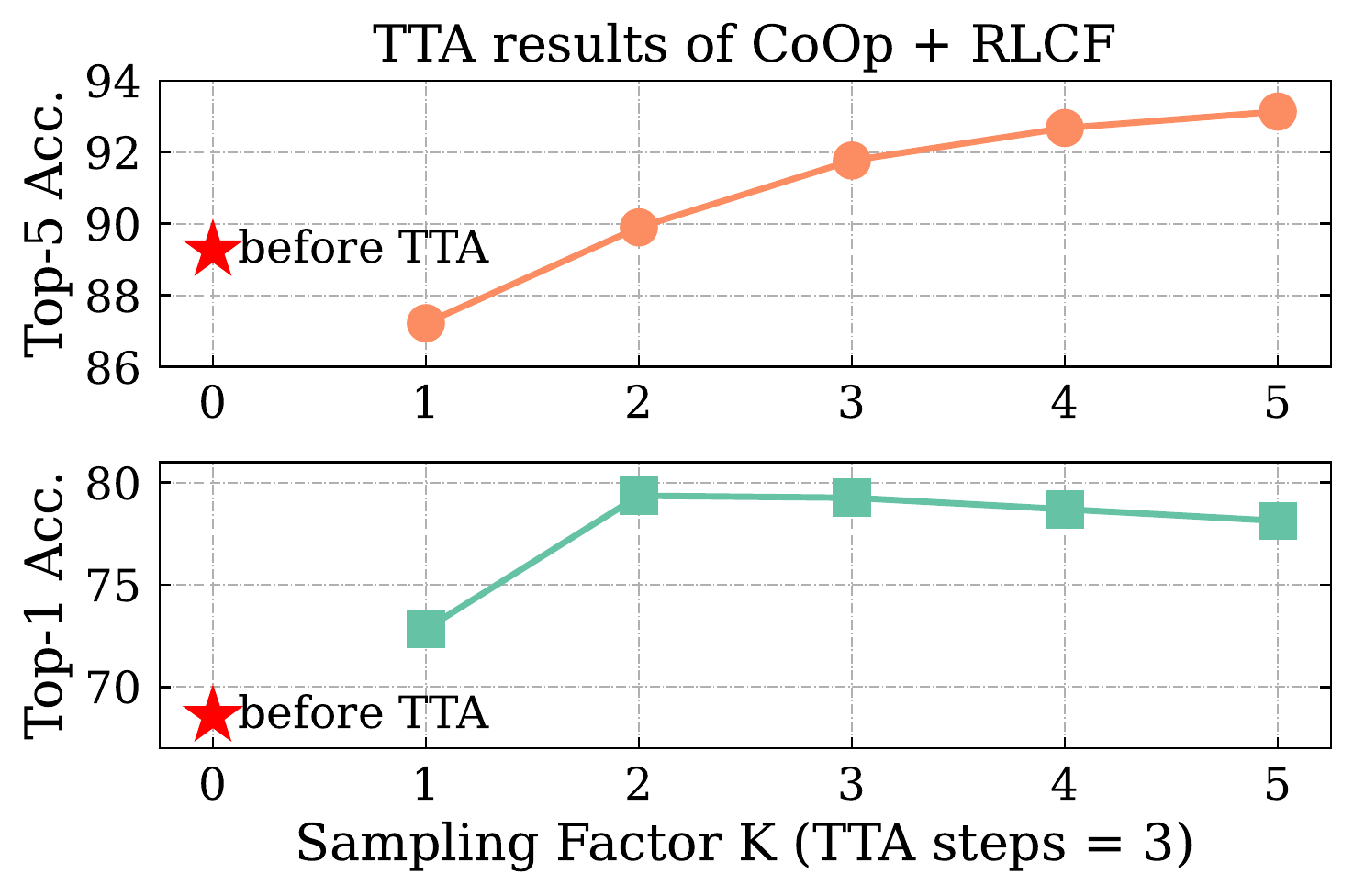}
\caption{\textbf{Different sampling factors $K$ in image classification on OOD data}.
 }
\label{fig:cls-k}
\end{wrapfigure}

When using RLCF with TTA for various VL tasks, we sample $K$ candidates from output distribution for reward calculations. In this section, we examine how $K$ affects different tasks and models.

\paragraph{Image classification on OOD data}
Figure~\ref{fig:cls-k} shows the average top-1 and top-5 accuracy on ImageNet-A, ImageNet-V2, and ImageNet-R for different $K$ in image classification.
RLCF reduces to pseudo-label~\citep{lee2013pseudo} when $K=1$.
A larger $K$ improves top-5 accuracy, but not top-1.
Too many sampled classes may make the optimization process difficult for policy gradient.
For example, when $K=5$ and only one class gets a positive score and other classes get negative scores, pushing away 4 negative classes may cause unpredictable behavior and make the model miss the ground truth after gradient updating.

\begin{table}[ht]
\vspace{-0.5cm}
	\caption
	{
\textbf{Different sampling factors $K$ in zero-shot text-image retrieval}.
$K_{t2i}$ for sampling in text-to-image retrieval,
and $K_{i2t}$ for sampling in image-to-text retrieval.
	}
	\centering	
    \setlength\tabcolsep{4pt}
	\resizebox{\textwidth}{!}{%
	\begin{tabular}	{cc|cccccc|cccccc}
		\toprule	 	
    \multicolumn{2}{c}{\multirow{3}{*}{Method}} &
    \multicolumn{6}{c|}{MS-COCO (5K test images)} & \multicolumn{6}{c}{Flickr30K (1K test images)} \\

    &
    & \multicolumn{3}{c}{text-to-image}
    & \multicolumn{3}{c|}{image-to-text}
    & \multicolumn{3}{c}{text-to-image}
    & \multicolumn{3}{c}{image-to-text} \\

    &
    & R@1 &R@5 &R@10
    & R@1 &R@5 &R@10
    & R@1 &R@5 &R@10
    & R@1 &R@5 &R@10\\
    \midrule
    & & \multicolumn{12}{c}{\textit{\underline{Zero-shot baseline}}}  \\

    \multicolumn{2}{c}{CLIP-ViT-B/16}
    & 33.0	& 58.2	& 68.9
    & 52.5	& 76.8	& 84.6
    & 62.2	& 85.7	& 91.8
    & 81.2	& 96.4	& 98.5	\\

    \midrule
    $K_{t2i}$ & $K_{i2t}$ &
    \multicolumn{12}{c}{\underline{\textit{TTA for }CLIP-ViT-B/16 with \textbf{RLCF}}}  \\

    6 & 14
    & \cellcolor{Gray1}37.8 & 61.1 & 68.1
    & 59.1 & 80.3 & 86.5
    & 66.6
    & 87.9
    & 91.5
    & 86.7
    & 97.6
    & 98.6	\\

    8 & 16
    & 37.6 & 62.1 & 69.6
    & 59.3 & 80.4 & 86.8
    & 66.7
    & 89.0
    & 92.1
    & 87.3
    & 97.2
    & \cellcolor{Gray1}98.8	\\

    10 & 18
    & 37.4 & 62.4 & 70.8
    & 59.5 & 80.1 & 86.9
    & 67.0 & 89.0 & 92.7
    & 87.1 & 97.2 & 98.7	\\

    12 & 20
    & 37.3
    & \cellcolor{Gray1}62.7	
    & 71.5
    & 59.1
    & 80.1
    & 86.9
    & \cellcolor{Gray1}67.1 
    & \cellcolor{Gray1}89.1 & 93.2
    & 87.2 & \cellcolor{Gray1}97.3 & 98.6 \\

    14 & 22
    & 37.0 & 62.5 & 72.4
    & \cellcolor{Gray1}59.8 & 80.2 & 86.8
    & 66.9 & 89.0 & 93.4
    & 87.3 & \cellcolor{Gray1}97.3 & 98.6	\\

    16 & 24
    & 36.8 & 62.4 & \cellcolor{Gray1}72.5
    & 59.5 
    & \cellcolor{Gray1}80.5 & \cellcolor{Gray1}87.3
    & 66.9 & 88.8 & \cellcolor{Gray1}93.5
    & \cellcolor{Gray1}87.6 & 96.9 & 98.7	\\

    18 & 26
    & 36.9 & 62.5 & \cellcolor{Gray1}72.5
    & 59.2 & 80.3 & 87.0
    & 66.7
    & 88.8
    & \cellcolor{Gray1}93.5
    & 87.3 & 96.8 & 98.6	\\

	\bottomrule
	\end{tabular}}
    \label{tab:retrieval-k}
\end{table}		

\paragraph{Zero-shot text-image retrieval}
Table~\ref{tab:retrieval-k} presents the effect of sampling factor $K$ in zero-shot text-image retrieval.
Similar to image classification, a larger $K$ generally leads to better Recall@5 and Recall@10 compared to a smaller $K$.
However, a smaller $K$ tends to produce better Recall@1 in most cases.
In MS-COCO and Flickr30K, one image has 5 reference captions, so the sampling factor for image-to-text retrieval is larger than $K$ for text-to-image retrieval.

\paragraph{Image captioning}

Table~\ref{tab:caption-k} illustrates the effect of different values of sampling factor $K$ in cross-domain image captioning.
The optimal $K$ varies for different image captioning models.
CLIPCap has better captioning capabilities than CapDec, so it can produce better candidates. Therefore, a larger $K$ is suitable for CLIPCap.

From the ablation study of sampling factor $K$, we find that the choice of $K$ depends on the tasks and VLMs. Different tasks and models require various sampling strategies.


\begin{table}[t]
    \centering
    \caption{\textbf{Different sampling factors $K$ in image captioning}.
    Metrics B@4 for BLEU@4, C for CIDEr, S for SPICE,
    and Ref-C for RefCLIPScore.
    }
    \setlength\tabcolsep{4pt}
    \resizebox{\textwidth}{!}{%
    \begin{tabular}{l|ccccccccc|ccccc}
    \toprule
    \multirow{3}{*}{Method}
    & \multicolumn{9}{c|}{MS-COCO $\Longrightarrow$ NoCaps }
    & \multicolumn{4}{c}{MS-COCO $\Longrightarrow$ Flickr30K } \\

    & \multicolumn{3}{c}{in domain}
    & \multicolumn{3}{c}{near domain}
    & \multicolumn{3}{c|}{out domain}
    & \multicolumn{4}{c}{Karpathy's test split} \\

    & B@4 & C & S
    & B@4 & C & S
    & B@4 & C & S
    & B@4 & C & S & Ref-C \\
    \midrule

    & \multicolumn{13}{c}{\underline{\textit{TTA for} CapDec~(\textbf{zero-shot})}} \\

    CapDec~\citep{nukrai2022text}
    & 32.4 & 62.6 & 10.3
    & 29.2 & 54.0 & 9.6
    & 17.2 & 31.7 & 6.4
    & 19.3 & 37.0 & 11.7 & 74.1 \\

    + \textbf{RLCF}~($K=4$)
    & 32.7 & 65.5 
    & \cellcolor{Gray1}10.7
    & 30.0 & 57.5 & 10.2
    & 17.1 & 34.6 & 6.8
    & 20.2
    & 40.8
    & 12.5
    & 75.4 \\

    + \textbf{RLCF}~($K=6$)
    & \cellcolor{Gray1}33.3 
    & \cellcolor{Gray1}68.0 
    & \cellcolor{Gray1}10.7
    & \cellcolor{Gray1}30.3 
    & \cellcolor{Gray1}57.9
    & \cellcolor{Gray1}10.3
    & \cellcolor{Gray1}17.6 
    & \cellcolor{Gray1}35.5
    & \cellcolor{Gray1}6.9
    & \cellcolor{Gray1}20.3
    & \cellcolor{Gray1}41.9
    & 12.7
    & 75.7 \\

    + \textbf{RLCF}~($K=8$)
    & 32.7 & 65.5 
    & \cellcolor{Gray1}10.7
    & 30.0 & 57.5 & 10.2
    & 17.1 & 34.6 & 6.8
    & 20.2 & 41.8 
    & \cellcolor{Gray1}12.9
    & \cellcolor{Gray1}75.8\\

    \midrule
    & \multicolumn{13}{c}{\underline{\textit{TTA for} CLIPCap~(\textbf{cross-domain})}} \\

    CLIPCap~\citep{mokady2021clipcap}
    & 36.3 & 76.9 & 11.9
    & 34.8 & 73.5 & 11.0
    & 22.5 & 54.6 & 8.6
    & 21.8 & 49.3 & 13.1 & 76.7 \\

    + \textbf{RLCF}~($K=8$)
    & 38.4 & 83.3 
    & \cellcolor{Gray1}12.5
    & 36.0 & 78.5 & 11.6
    & 24.1 & 61.4 & 9.3
    & 22.7
    & 55.8
    & 14.3
    & 79.3 \\

    + \textbf{RLCF}~($K=10$)
    & \cellcolor{Gray1}38.6 
    & \cellcolor{Gray1}84.0
    & \cellcolor{Gray1}12.5
    & \cellcolor{Gray1}36.1 
    & 79.6
    & 11.8
    & \cellcolor{Gray1}24.7
    & \cellcolor{Gray1}63.8
    & \cellcolor{Gray1}9.6
    & \cellcolor{Gray1}23.3 
    & 56.6
    & 14.5 
    & \cellcolor{Gray1}79.4 \\

    + \textbf{RLCF}~($K=12$)
    & 38.5 & 82.0 
    & \cellcolor{Gray1}12.5
    & 35.9 
    & \cellcolor{Gray1}80.2 
    & \cellcolor{Gray1}11.9
    & 24.4 & 63.1 & 9.5
    & 23.1
    & \cellcolor{Gray1}57.5
    & \cellcolor{Gray1}14.6
    & \cellcolor{Gray1}79.4 \\

    \bottomrule
    \end{tabular}}
\label{tab:caption-k} 
\vspace{-0.2cm}
\end{table}

\subsection{Different Reward Models}
RLCF relies on the good quality of the CLIP reward models.
In this section, we show the influence of different CLIP reward models in image classification and image captioning.

As shown in Table~\ref{tab:diff-reward-cls}, RLCF is robust to different reward models. Compared to the baseline CoOp, RLCF can achieve improvements even with a CLIP-RN50$\times$4 as the reward model, which is worse than the prompt tuning model CLIP-ViT-B/16.
When the prompt tuning model and the reward model are the same, RLCF is also better than the state-of-the-art test-time prompt tuning method --- TPT~\citep{shu2022tpt}.
With CLIP-ViT-L/14 as the reward model, RLCF with prompt tuning is slightly better than the ensemble result. It is worth noting that RLCF with image encoder tuning in Table~\ref{tab:ood-main} is obviously better than the ensemble results in the OOD average performance.
Compared to the ensemble method, RLCF can adapt to the test distribution with the feedback mechanism.
This is why RLCF shows better performance than the ensemble results.

We also test different reward models in image captioning. Results are shown in Table~\ref{tab:diff-reward-cap}.
CapDec and CLIPCap both use CLIP-ViT-B/16 as the image embedding extractor.
RLCF with different reward models can always achieve significant improvements in the near and out domain data.
This shows the robustness of RLCF to open domain scenarios with different CLIP reward models.

\begin{table}[!t]
\vspace{-0.5cm}
  \caption{\textbf{RLCF with different reward models in image classification}.
  "--" for not available.}
  \label{tab:diff-reward-cls}
  \centering
  \resizebox{\linewidth}{!}{%
  \begin{tabular}{l|c|*{5}c}
    \toprule
    Method & Reward Model
    &  ImageNet-A 
    & ImageNet-V2       & ImageNet-R
    & ImageNet-Sketch   & OOD Average \\
    \midrule

    & \multicolumn{6}{c}{\textit{\underline{Zero-shot baseline}}}  \\
    CLIP-RN50$\times$4
    & --
    & 39.66  &   58.72    
    &  69.10 &  42.73  &   52.55  \\
    CLIP-ViT-B/16  
    & --
    & 47.87  &   60.86    
    &  73.98 &  46.09  &   57.20  \\
    CLIP-ViT-L/14  
    & --
    &  68.82  &   67.80    
    &  85.40  &  57.84  &   69.97  \\
    Ensemble~(B/16 + L/14) 
    & --
    &  65.94  &   69.02    
    &  85.92 &  57.98  &   69.72  \\
  
    \midrule
    & \multicolumn{6}{c}{\underline{\textit{Prompt tuning for} CLIP-ViT-B/16}} \\
    CoOp~\citep{zhou2021coop}
    & --
    &  49.71  &   64.20
    &  75.21  &  47.99  &   59.28  \\

    TPT~+~CoOp~(\citeauthor{shu2022tpt})    
    & --  
    & 57.95  & 66.83
    & 77.27  & 49.29  & 62.84 \\

    RLCF~+~CoOP
    & CLIP-RN50$\times$4
    &  52.06  &   64.62
    &  76.00  &  48.42  &   60.28  \\
    RLCF~+~CoOP
    & CLIP-ViT-B/16
    &  61.66  &   67.04
    &  78.06  &  49.70  &   64.12  \\
    RLCF~+~CoOP
    & CLIP-ViT-L/14
    &  \cellcolor{Gray1}{69.74}  
    &   \cellcolor{Gray1}{70.62}
    &  \cellcolor{Gray1}{84.51}  
    &  \cellcolor{Gray1}{56.49}  
    &   \cellcolor{Gray1}{70.34}  \\
    \bottomrule
  \end{tabular}
}
\end{table}


\begin{table}[t]
    \centering
    \caption{\textbf{RLCF with different reward models in image captioning}.
    Metrics M for METEOR, C for CIDEr, S for SPICE,
    and Ref-C for RefCLIPScore. We keep the embedding extractor for the two methods as CLIP-ViT-B/16.
    }
    \label{tab:diff-reward-cap} 
    \setlength\tabcolsep{4pt}
    \resizebox{\textwidth}{!}{%
    \begin{tabular}{l|c|c|cccc|cccc|cccc}
    \toprule
    \multirow{3}{*}{Method}
    & Embed. Extractor
    & Reward Model
    & \multicolumn{12}{c}{MS-COCO $\Longrightarrow$ NoCaps } \\
    
    & &
    & \multicolumn{4}{c|}{in domain}
    & \multicolumn{4}{c|}{near domain}
    & \multicolumn{4}{c}{out domain} \\

    & &
    & M & C & S & Ref-C
    & M & C & S & Ref-C
    & M & C & S & Ref-C \\
    \midrule

    & &
    & \multicolumn{12}{c}{\underline{\textit{TTA for} CapDec~\citep{nukrai2022text}~(\textbf{zero-shot})}} \\

    CapDec
    & CLIP-ViT-B/16 
    & --
    & 23.9 & 62.6 & 10.3 & 75.5
    & 22.3 & 54.0 & 9.6 & 74.2 
    & 17.2 & 31.7 & 6.4 & 71.3 \\
    
     + \textbf{RLCF}
    & CLIP-ViT-B/16 
    & CLIP-ViT-B/16 
    & 24.3 & 65.5 & 10.7 & 76.6
    & 22.7 & 57.2 & 10.0 & 75.7 
    & 17.7 & 35.0 & 6.8 
    & \cellcolor{Gray1}72.9 \\    

     + \textbf{RLCF}
    & CLIP-ViT-B/16 
    & CLIP-ViT-L/14
    & \cellcolor{Gray1}24.6 
    & \cellcolor{Gray1}68.0 
    & \cellcolor{Gray1}10.7 
    & \cellcolor{Gray1}76.7
    & \cellcolor{Gray1}23.0 
    & \cellcolor{Gray1}57.9 
    & \cellcolor{Gray1}10.3 
    & \cellcolor{Gray1}75.7 
    & \cellcolor{Gray1}17.9 
    & \cellcolor{Gray1}35.5 
    & \cellcolor{Gray1}6.9 
    & 72.7 \\


    \midrule
    & &
    & \multicolumn{12}{c}{\underline{\textit{TTA for} CLIPCap~\citep{mokady2021clipcap}~(\textbf{cross-domain})}} \\

    CLIPCap
    & CLIP-ViT-B/16
    & --
    & 26.4 & 76.9 & 11.9 & 77.8
    & 24.8 & 73.5 & 11.0 & 77.6
    & 20.3 & 54.6 & 8.6 & 75.7 \\

     + \textbf{RLCF}
    & CLIP-ViT-B/16 
    & CLIP-ViT-B/16 
    & 26.9 & 81.1 & 12.3 & 80.1
    & 25.5 & 78.1 & 11.7 & 80.1
    & 21.2 & 62.0 & 9.5  & 78.5 \\    

     + \textbf{RLCF}
    & CLIP-ViT-B/16 
    & CLIP-ViT-L/14
    & \cellcolor{Gray1}27.2
    & \cellcolor{Gray1}84.0
    & \cellcolor{Gray1}12.5
    & \cellcolor{Gray1}80.3
    & \cellcolor{Gray1}25.7 
    & \cellcolor{Gray1}79.6 
    & \cellcolor{Gray1}11.8 
    & \cellcolor{Gray1}80.1 
    & \cellcolor{Gray1}21.5 
    & \cellcolor{Gray1}63.8 
    & \cellcolor{Gray1}9.6 
    & \cellcolor{Gray1}78.5 \\

    \bottomrule
    \end{tabular}}

\end{table}

\subsection{GPU runtime and memory}
\begin{table}[!t]
	\caption
	{
\textbf{Average GPU inference time per sample and GPU memory with different TTA steps}. Test on ImageNet-A and ImageNet-V2 with a single NVIDIA 40GB A100 GPU.
	}
	\centering	
	\resizebox{\textwidth}{!}{%
	\begin{tabular}	{l|c|ccc|ccc}
	\toprule	 	
    CLIP-ViT-B/16 & TTA Steps &
    \multicolumn{3}{c|}{ImageNet-A} &
    \multicolumn{3}{c}{ImageNet-V2} \\
    \midrule

    \textbf{Prompt Tuning} &
    & Acc.  & Mem. (GB) & Time (s) 
    & Acc.  & Mem. (GB) & Time (s)   \\

    TPT~+~CoOp & 1 &
    57.95	& 4.2	& 0.168	&
    66.83	& 18.2	& 0.468 \\

    TPT~+~CoOp & 3 &
    60.13	& 4.2	& 0.320	&
    66.76	& 18.2	& 1.05 \\

    RLCF~+~CoOp & 1 &
    63.07	& 6.2	& 0.197	&
    69.59	& 19.8	& 0.486 \\

    RLCF~+~CoOp & 3 &
    \textbf{69.74}	& 6.2	& 0.348	&
    \textbf{70.62}	& 19.8	& 1.08 \\

    \midrule

    \textbf{Image encoder tuning} &
    & Acc.  & Mem. (GB) & Time (s) 
    & Acc.  & Mem. (GB) & Time (s)   \\

    TPT~+~CoOp & 1 &
    61.78	& 8.8	& 0.208	&
    63.70	& 8.8	& 0.272 \\

    TPT~+~CoOp & 3 &
    62.07	& 8.8	& 0.384	&
    64.02	& 8.8	& 0.512 \\

    RLCF & 1 &
    71.23	& 10.8	& 0.239	&
    67.60	& 10.8	& 0.319 \\

    RLCF & 3 &
    \textbf{73.71}	& 10.8	& 0.415	&
    \textbf{69.77}	& 10.8	& 0.543 \\

	\bottomrule
	\end{tabular}}
    \label{tab:gputime}
\vspace{-0.2cm}
\end{table}

Efficiency is also important in TTA.
We provide the GPU time and memory in Table~\ref{tab:gputime}.

\textbf{Compared to TPT, the inference time of RLCF increases by a constant amount for different TTA steps and datasets, \ie, roughly 0.03s per sample}.
For each sample, the CLIP reward model only needs to run the image encoder once. This is the source of the 0.03s increase. The text features of CLIP reward model are always the same because the class names are fixed.

ImageNet-A has 200 classes, while ImageNet-V2 has 1000 classes. For prompt tuning on ImageNet-V2, the input batch size of the CLIP text encoder is 1000, and we need to re-run the text encoder to update the text features after each TTA step. This is why prompt tuning is slower and consumes more memory than image encoder tuning. For image encoder tuning, the text features are unchanged and the image encoder only has a single image as input.

\end{document}